%% file: main.tex
\documentclass[10pt,twocolumn,letterpaper]{article}

\usepackage[pagenumbers]{iccv} 
\input{preamble}

\definecolor{iccvblue}{rgb}{0.21,0.49,0.74}
\usepackage{subfiles}
\usepackage[pagebackref,breaklinks,colorlinks,allcolors=iccvblue]{hyperref}
\def\paperID{2670} 
\def\confName{ICCV}
\def\confYear{2025}

\title{Flux-Sculptor: Text-Driven Rich-Attribute Portrait Editing through Decomposed Spatial Flow Control}

\author{
  Tianyao He\textsuperscript{1,2} \ \ 
  Runqi Wang\textsuperscript{1} \ \ 
  Yang Chen\textsuperscript{1} \ \ 
  Dejia Song\textsuperscript{1} \ \ 
  Nemo Chen\textsuperscript{1} \ \ 
  Xu Tang\textsuperscript{1} \ \ 
  Yao Hu\textsuperscript{1} \\
  \textsuperscript{1}Xiaohongshu \quad
  \textsuperscript{2}Shanghai Jiao Tong University \\
  \textbf{Project: \href{https://flux-sculptor.github.io/}{https://flux-sculptor.github.io/}}
}

\begin{document}
\maketitle

\begin{strip}\centering
\includegraphics[width=0.9\textwidth]{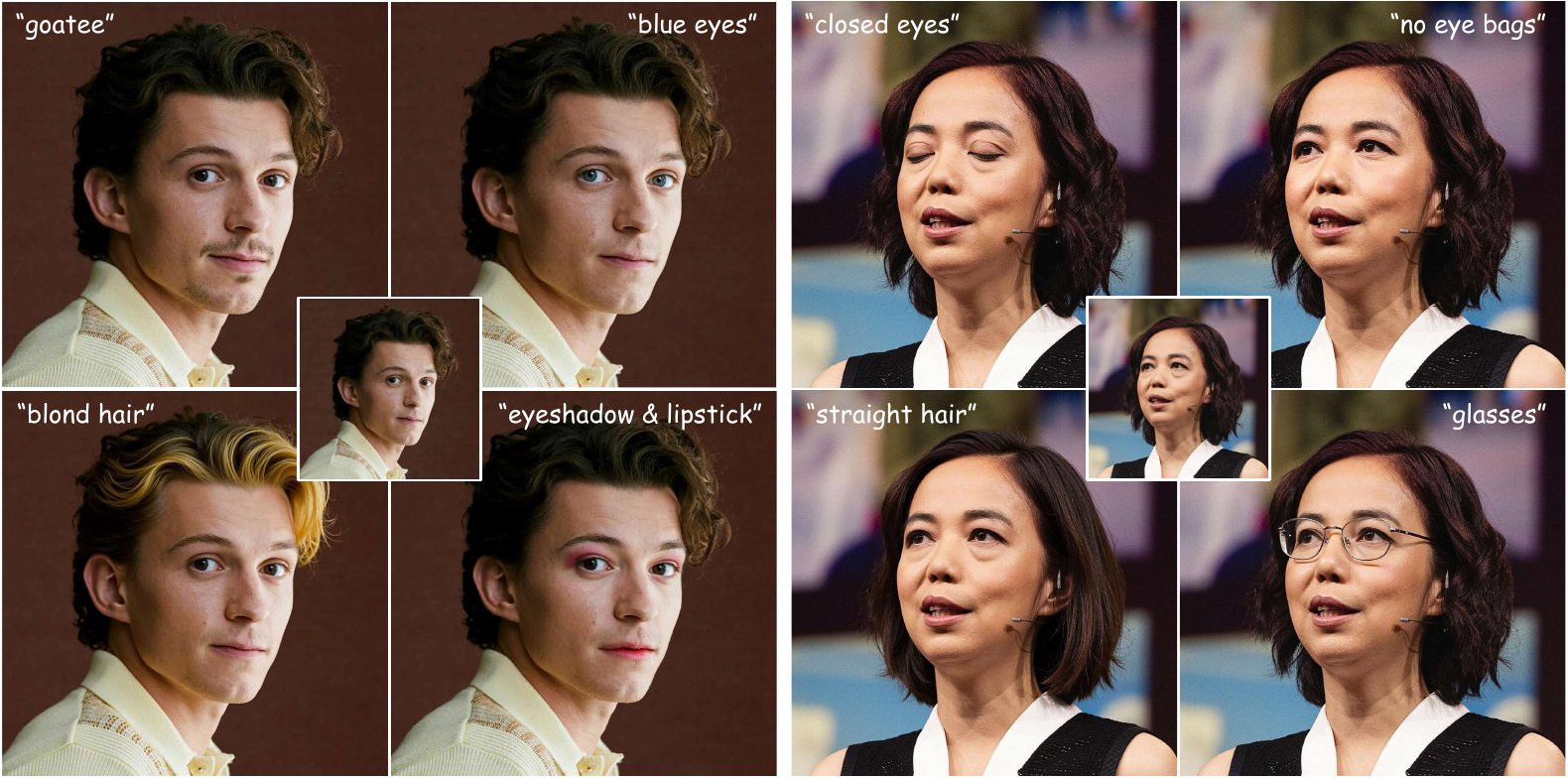}
\captionof{figure}{\textbf{Text-based portrait editing results of Flux-Sculptor.} We use eight text prompts to edit the central portraits. Flux-Sculptor effectively follows the text to achieve diverse attribute modification with visual harmony, naturalness, and identity preservation.}
\label{fig:teaser}

\end{strip}

\begin{abstract}
    Text-driven portrait editing holds significant potential for various applications but also presents considerable challenges. An ideal text-driven portrait editing approach should achieve precise localization and appropriate content modification, yet existing methods struggle to balance reconstruction fidelity and editing flexibility. To address this issue, we propose Flux-Sculptor, a flux-based framework designed for precise text-driven portrait editing. Our framework introduces a Prompt-Aligned Spatial Locator (PASL) to accurately identify relevant editing regions and a Structure-to-Detail Edit Control (S2D-EC) strategy to spatially guide the denoising process through sequential mask-guided fusion of latent representations and attention values. Extensive experiments demonstrate that Flux-Sculptor surpasses existing methods in rich-attribute editing and facial information preservation, making it a strong candidate for practical portrait editing applications. Project page is available at \href{https://flux-sculptor.github.io/}{https://flux-sculptor.github.io/}.
\end{abstract}
\vspace{-0.4cm}
\section{Introduction}
\label{sec:intro}
    Portrait editing technology empowers individuals to enhance their appearance, refine facial details, and modify stylistic elements to better align with their aesthetic preferences~\cite{pang2023dpe,gu2019mask,gao2021high}. In the era of social media, portrait editing has become an integral part of digital interactions, serving purposes such as personal branding, artistic expression, and online presence enhancement~\cite{chen2019association}. As a result, high-quality portrait editing plays a crucial role in self-representation and content creation, enabling individuals to present themselves in the most favorable manner.
    
    Compared to general image editing~\cite{kawar2023imagic,brooks2023instructpix2pix}, portrait editing demands higher precision, naturalness, and aesthetic appeal to meet user expectations. However, most existing editing tools~\cite{photoshop,lightroom} require a certain level of image manipulation expertise, making it difficult for casual users to achieve satisfactory results. This challenge highlights the need for a more intuitive portrait editing approach—one that enables text-based modifications while providing fine-grained control over the edits.

    Currently, the core technical challenge in text-based portrait editing is the trade-off between reconstruction fidelity and editing flexibility. While existing methods have made notable progress~\cite{yue2023chatface,jiang2021talk}, they still struggle to fully meet user expectations in terms of both accuracy and consistency. This raises a fundamental question: \textit{What defines an ideal text-based portrait edit?} In this paper, we break this down into two key criteria: \textit{precise editing localization} and \textit{appropriate content modification}.
    \begin{itemize}
        \item \textbf{Precise editing localization:} Edits should be strictly confined to the regions specified by the text prompt, ensuring that unrelated areas retain their original appearance.
        \item \textbf{Appropriate content modification:} Edits should accurately reflect the textual prompt while maintaining the natural coherence of the entire portrait.
    \end{itemize}
    These two criteria ensure that edits not only align with user expectations but also preserve visual harmony, naturalness, and identity consistency. They serve as fundamental principles guiding our method design and evaluation.

    However, current image editing methods still face substantial challenges in meeting the ideal performance criteria. Early GAN-based approaches~\cite{jiang2021talk,patashnik2021styleclip,wang2024maniclip} demonstrate efficacy in preserving subject identity and structural integrity, ensuring edited portraits maintain visual consistency with the original input. Nevertheless, these methods are constrained by limited generative flexibility and inadequate text comprehension capabilities, which impedes their ability to implement complex attribute modifications driven by textual instructions. Conversely, probabilistic flow models trained on large-scale text-to-image datasets~\cite{rombach2022high,liu2022flow} have demonstrated remarkable advancements in textual understanding and generative diversity. These models can produce modifications that accurately correspond to a wide spectrum of user-provided textual prompts. However, the inherent stochasticity of their denoising process frequently results in the deterioration of fine-grained details, unintended alterations to non-target regions, and even significant identity distortions—outcomes that are fundamentally unacceptable in portrait editing applications.
    
    To achieve ideal text-driven, rich-attribute portrait editing, we propose Flux-Sculptor, a novel framework based on decomposed spatial flow control. This framework follows an inversion-and-editing paradigm and spatially guides the denoising process to simultaneously ensure precise editing localization and appropriate content modification.
    To achieve precise editing localization, we design the Prompt-Aligned Spatial Locator (PASL) to identify the regions of a portrait relevant to the user-provided text prompt. PASL effectively transforms the user's textual prompt into edit and reconstruction regions for subsequent spatial control.
    For appropriate content modification, we further explore the temporal relationship between denoising and portrait editing. Observations reveal that the initial denoising steps establish the structural outline of the portrait, while later steps refine the texture and details. To address this, we design the Structure-to-Detail Edit Control (S2D-EC) strategy, which decomposes the editing flow into two stages: facial structuring and detailing. The original portrait information in the reconstruction regions is injected into these two stages via latent masking and attention masking strategies, respectively. Our framework fully utilizes the generative capabilities of the probabilistic flow model within a controllable range, enabling rich-attribute and highly detailed text-based portrait editing with enhanced facial attributes, as illustrated in Figure~\ref{fig:teaser}. In conclusion, our contributions are three-fold:
    \begin{itemize}
        \item We propose a decomposed spatial flow control framework that utilizes rectified flow for text-based, rich-attribute portrait editing, achieving precise localization and appropriate content.
        \item Within this framework, we first design the Prompt-Aligned Spatial Locator (PASL) to identify regions for editing based on the text prompt. Next, we introduce the Structure-to-Detail Edit Control (S2D-EC) strategy to ensure editing results align with the text prompt while maintaining strong natural consistency.
        \item Through extensive experiments, our framework outperforms existing portrait/image editing methods in terms of editing quality, identity and detail preservation, and the richness of editable attributes.
    \end{itemize}

\section{Related Work}
\subsection{Text-Driven Image Editing}
    Text-driven image editing has advanced alongside developments in probabilistic flow-based models, such as diffusion~\cite{ho2020denoising, song2020denoising} and rectified flow~\cite{liu2022flow}.
    To address the lack of editing data, InstructPix2Pix~\cite{brooks2023instructpix2pix} synthesized an instruction-image pair dataset using GPT-3~\cite{brown2020language} and Prompt-to-Prompt~\cite{hertz2022prompt}, training an instruction-guided image editing diffusion model. To mitigate synthesized noise, MagicBrush~\cite{zhang2023magicbrush} manually annotated an instruction-guided real image editing dataset and improved InstructPix2Pix through fine-tuning. However, supervision based on edited pairs cannot guarantee consistent editing, leading to inevitable loss of detail.
    With the advent of large-scale text-to-image generative models (e.g., SD1.5, Flux), researchers have started exploring the training-free inversion-and-editing paradigm. Among these, MasaCtrl~\cite{cao2023masactrl} first uses DDIM inversion, followed by a mask-guided mutual self-attention strategy to control the denoising process. However, its coarse foreground masks and control strategy fail to achieve fine-grained localization, limiting its effectiveness for rich-attribute portrait editing.
    Recent works have begun to explore rectified flow in image editing. RF-inversion~\cite{rout2024semantic} uses dynamic optimal control to perform inversion and editing, while RF-Edit~\cite{wang2024taming} improves inversion and editing accuracy through a high-order solver and self-attention feature sharing. Despite their visually appealing results, these methods still introduce noticeable, unrelated changes due to the lack of spatial constraints.
    
\subsection{Portrait Editing}
    Portrait editing aims to naturally modify the required facial attributes without altering identities or details. GAN-based methods achieve portrait editing by modifying the latent space through StyleGAN inversion. Among these, Talk-to-Edit~\cite{jiang2021talk}, StyleCLIP~\cite{patashnik2021styleclip}, and ManiCLIP~\cite{wang2024maniclip} use language encoders to interpret text requirements and adjust the image latent in the semantic space. These methods can produce consistent editing results, but their editable attributes and text comprehension capabilities are highly dependent on the training process, which limits their generative flexibility and diversity.
    For diffusion-based portrait editing, DAE~\cite{preechakul2022diffusion} decomposes high-level semantics and stochastic details, enabling attribute modification through classifier guidance. Additionally, ChatFace~\cite{yue2023chatface} adapts ChatGPT to DAE for interpreting user requests and translating them into target attributes and editing guidance. However, these methods require precomputing and recording modification directions for each attribute using trained linear classifiers, making them complex and difficult to handle open-set text requirements.
    Furthermore, there are methods with excellent performance but limited to specific facial domains, such as facial blemish modification~\cite{jiang2024hunting}, hairstyle editing~\cite{zhang2024stable}, and makeup enhancement~\cite{jin2024toward, li2018beautygan}. Unlike these existing works, our approach aims to overcome current limitations in comprehension and editing capabilities, enabling text-driven, rich-attribute portrait editing.
    
\begin{figure*}
    \centering
    \includegraphics[width=0.9\linewidth]{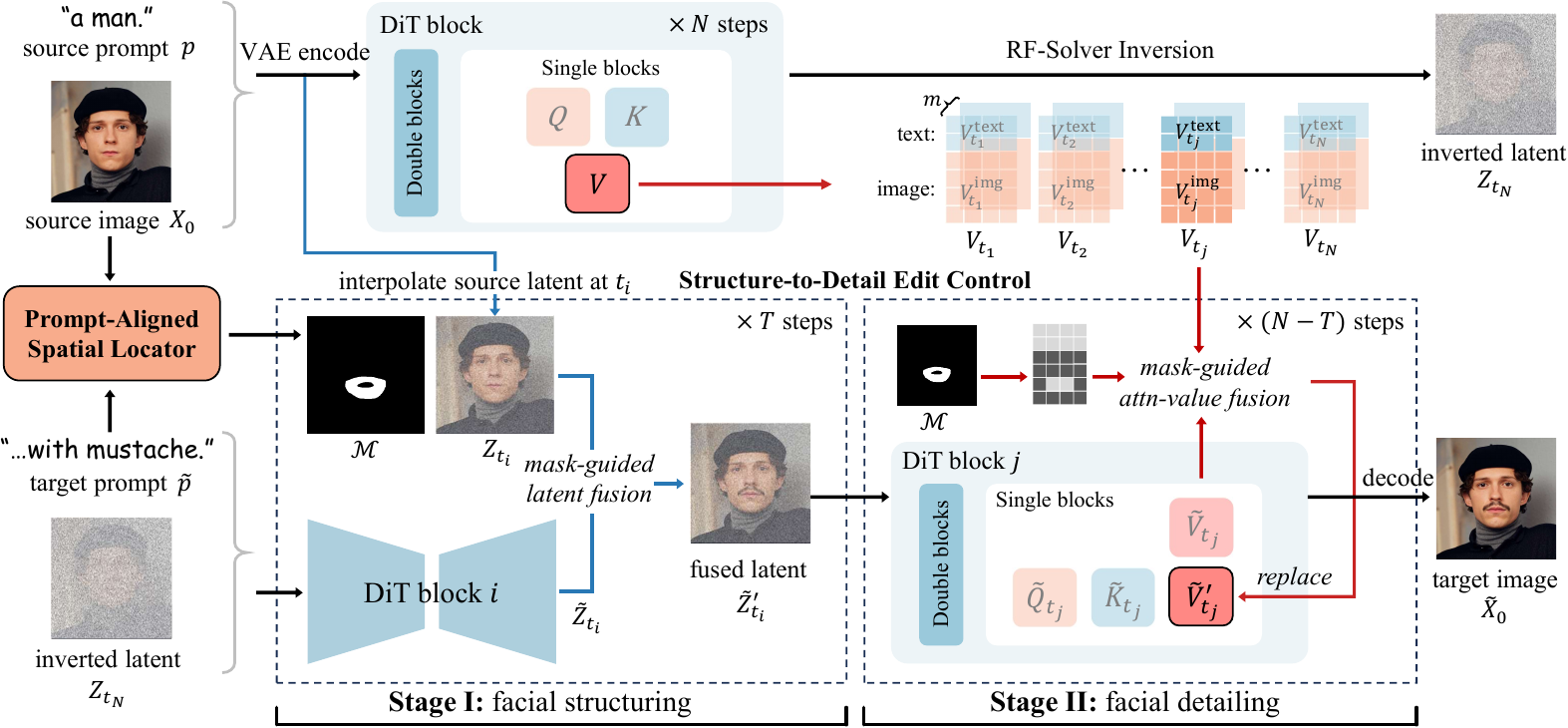}
    \caption{\textbf{The Flux-Sculptor pipeline.} Flux-Sculptor first obtains the inverted latent through RF-Solver Inversion. Then, the PASL extracts the target text-related editing regions based on the source image and target prompt. Based on the editing regions, the framework designs the S2D-EC strategy to control facial structuring and detailing through region-guided latent fusion and attention-value fusion, respectively.}
    \label{fig:framework}
\end{figure*}
    
\section{Method}
\subsection{Preliminary: Rectified Flow}
    Rectified flow~\cite{liu2022flow} enables a smooth transition between the Gaussian noise distribution $\pi_0$ and the real data distribution $\pi_1$ along a straight path. This is achieved by simulating the forward process through an ordinary differential equation (ODE) with time $t\in[0,1]$:
    \vspace{-2pt}
    \begin{equation}
        \mathrm{d}Z_t=v(Z_t, t)\mathrm{d}t,\quad t\in [0,1],
        \label{eq:rf}
    \end{equation}
    where $Z_0\sim\pi_0$ and $Z_1\sim\pi_1$. In practice, the velocity function $v(\cdot)$ is parameterized by a neural network $v_{\theta}(Z_t, t)$. The forward process of rectified flow is formulated as a linear function with time $t\in[0,1]$:
    \begin{equation}
    \begin{aligned}
        Z_t &= tZ_1 + (1-t)Z_0 \\
        \xrightarrow[]{d/dt} \mathrm{d}Z_t &= (Z_1-Z_0)\mathrm{d}t,
    \end{aligned}
    \end{equation}
    which shows $(Z_1-Z_0)$ is the ground truth of $v_{\theta}(Z_t, t)$. We can optimize $v_{\theta}(Z_t, t)$ through least squares regression:
    \begin{equation}
        \min_{\theta}\int_{0}^1\mathbb{E}\left[ \lVert(Z_1-Z_0)-v_{\theta}(Z_t, t) \rVert^2 \right]\mathrm{d}t.
    \end{equation}
    The backward process starts at $Z_{t_N}\sim \mathcal{N}(0, I)$ and steps backward $N$ discreate timesteps $\{t_{N-1},\dots ,t_0\}$ according to the following the function:
    \begin{equation}
        Z_{t_{i-1}} = Z_{t_i} + (t_{i-1}-t_{i})v_{\theta}(Z_{t_i}, t_i).
    \end{equation}
    In conclusion, rectified flow defines a straightforward transformation path between noise and images, achieving more efficient generation compared to diffusion models.

\subsection{Overview}
    \noindent\textbf{Task formulation.}
    In the text-based protrait editing, the user provides a source image $X_0$, a source prompt $p$ and a target prompt $\tilde{p}$. The source prompt describes the source image, which can be simple (e.g., ``a man.''), while the target prompt describes the expected edited image (e.g., ``a man with mustache.''). Using three inputs, the model generates the edited image $\tilde{X_0}$ that aligns with the portrait editing criteria discussed in Section~\ref{sec:intro}.
    
    \noindent\textbf{Framework.}
    The pipeline of our proposed Flux-Sculptor is illustrated in Figure~\ref{fig:framework}. Our framework is established on the Flux model~\cite{liu2022flow} and follows an inversion-and-editing process. First, $X_0$ is encoded into latent $Z_{t_0}$ through VAE~\cite{kingma2013auto} and we use RF-Solver~\cite{wang2024taming} to inverse $Z_{t_0}$ to latent $Z_{t_N}\sim N(0, I)$. Then, for precise editing localization, the Prompt-Aligned Spatial Locator (PASL) is trained to extract the portrait regions relevant to the target prompt and generate the editing region mask $\mathcal{M}$. For appropriate content modification, we design a Structure-to-Detail Edit Control (S2D-EC) strategy to apply proper spatial control to the editing flow. The strategy decomposes the denoising process into a facial structuring stage and a detailing stage. In the early facial structuring stage, we use mask-guided latent fusion to inject the source image's structure information. In the later facial detailing stage, fusion is applied to attention values within the Diffusion Transformer (DiT) blocks~\cite{peebles2023scalable}, yielding the final edited image $\tilde{X_0}$. We will detail the PASL module and S2D-EC strategy in the following sections.

\begin{figure}[t]
    \centering
    \includegraphics[width=1.0\linewidth]{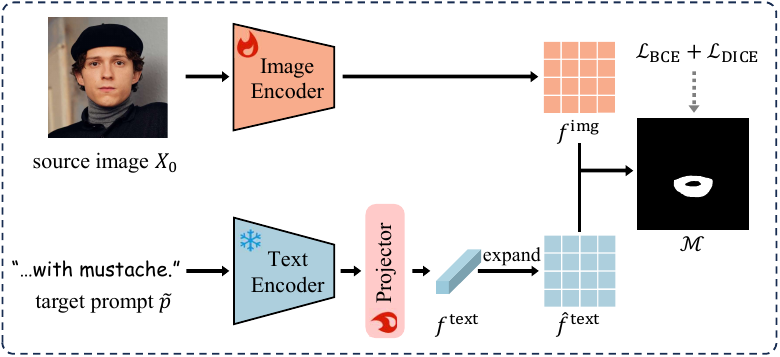}
    \caption{\textbf{The Prompt-Aligned Spatial Locator.} Given $X_0$ and $\tilde{p}$, the PASL extracts visual and language features and generates the editing mask based on cross-modal similarities.}
    \label{fig:PASL}
    \vspace{-0.2cm}
\end{figure}

\subsection{Prompt-Aligned Spatial Locator}
    As mentioned in Section~\ref{sec:intro}, the first criterion for ideal portrait editing is precise editing localization, which requires modifications to be strictly confined to the text-related regions, while the remaining parts should be reconstructed. Despite various open-set segmentation methods, existing segmenters fail to effectively locate facial attributes due to their complexity and fine granularity. Therefore, we design the Prompt-Aligned Spatial Locator (PASL) to locate the facial editing regions based on the target prompt.

    \noindent\textbf{Training data preparation.}
    We prepare the Text2Mask dataset to train our PASL with image, text prompts, and editing masks. Detailed descriptions of the dataset is provided in Section~\ref{sec:text2mask_dataset} and Supplementary Materials A.1.
    
    \noindent\textbf{Module design.}
    As illustrated in Figure~\ref{fig:PASL}, PASL consists of an image encoder, a text encoder and a projector. The image encoder $\mathrm{Enc}(\cdot)$ is based on convolutional neural network and is trained to extract distinguishable facial attribute features. For the text encoder, we adopt AlphaCLIP's text encoder~\cite{sun2024alpha} $\alpha\text{-}\text{CLIP}^{\mathrm{text}}(\cdot)$, as it is trained on large-scale image-text datasets with mask guidance. Its parameters are frozen to preserve AlphaCLIP's region-aware ability. We further design a trainable multi-modal projector $\text{Proj}(\cdot)$ to map AlphaCLIP's text feature into the visual feature space. Given the source image $X_0$ and target prompt $\tilde{p}$, we first extract the image feature $f^{\mathrm{img}}$ and text feature $f^{\mathrm{text}}$:
    \vspace{-2pt}
    \begin{equation}
        \begin{array}{ll}
            f^{\mathrm{img}} = \mathrm{Enc}(X_0), & f^{\mathrm{img}}\in\mathbb{R}^{H\times W\times C}, \\
            f^{\mathrm{text}} = \text{Proj}\left[\alpha\text{-}\text{CLIP}^{\mathrm{text}}(\tilde{p})\right], & f^{\mathrm{text}}\in \mathbb{R}^{C},
        \end{array}
    \end{equation}
    where $C$ is the dimension size. Then, we expand the shape of the text feature from $f^{\mathrm{text}}\in \mathbb{R}^{C}$ to $\hat{f}^{\mathrm{text}}\in \mathbb{R}^{H\times W\times C}$ through broadcasting, and flatten first two dimensions of output features. The first dimension $L=H\times W$ represents the total pixel number. The final editing mask $\mathcal{M}$ is generated by calculating the image-to-text similarity:
    \begin{equation}
        \mathcal{M}_{i} = \sigma\left(\sum_{c=1}^{C}\left[f^{\mathrm{img}}\circ \hat{f}^{\mathrm{text}}\right]_{i,c}\right),\  i\in[1,2,\dots,L],
    \end{equation}
    where $\circ$ represents the Hadamard product and $\sigma(\cdot)$ represents the sigmoid function.

    \noindent\textbf{Loss functions.}
    To supervised the predicted mask $\mathcal{M}$ with the ground truth mask $\mathcal{M}^{\text{gt}}$, we use the Binary Cross-Entropy (BCE) loss $\mathcal{L}_{\text{BCE}}$ and Dice Loss~\cite{sudre2017generalised} $\mathcal{L}_{\text{DICE}}$:
    \begin{equation}
            \mathcal{L}_{\text{BCE}} = - \frac{1}{L} \sum_{i=1}^{L} \left[ \mathcal{M}^{\text{gt}}_{i} \log(\mathcal{M}_i) + (1 - \mathcal{M}^{\text{gt}}_{i}) \log(1 - \mathcal{M}_i) \right]
    \end{equation}
    \begin{equation}
            \mathcal{L}_{\text{DICE}} = 1 - \frac{2 \sum_{i=1}^{L} \mathcal{M}_i \mathcal{M}^{\text{gt}}_{i}}{\sum_{i=1}^{L} \mathcal{M}_i + \sum_{i=1}^{L} \mathcal{M}^{\text{gt}}_{i}}.  
    \end{equation}
    Finally, the mask loss $\mathcal{L}_{\text{mask}}$ can be written as:
    \begin{equation}
        \mathcal{L}_{\text{mask}} = \mathcal{L}_{\text{BCE}} + \mathcal{L}_{\text{DICE}}.
    \end{equation}
    PASL can automatically generate precise facial editing regions based on the text prompts, which lays the foundation for ideal and user-friendly portrait editing.

\begin{figure}[t]
    \centering
    \includegraphics[width=\linewidth]{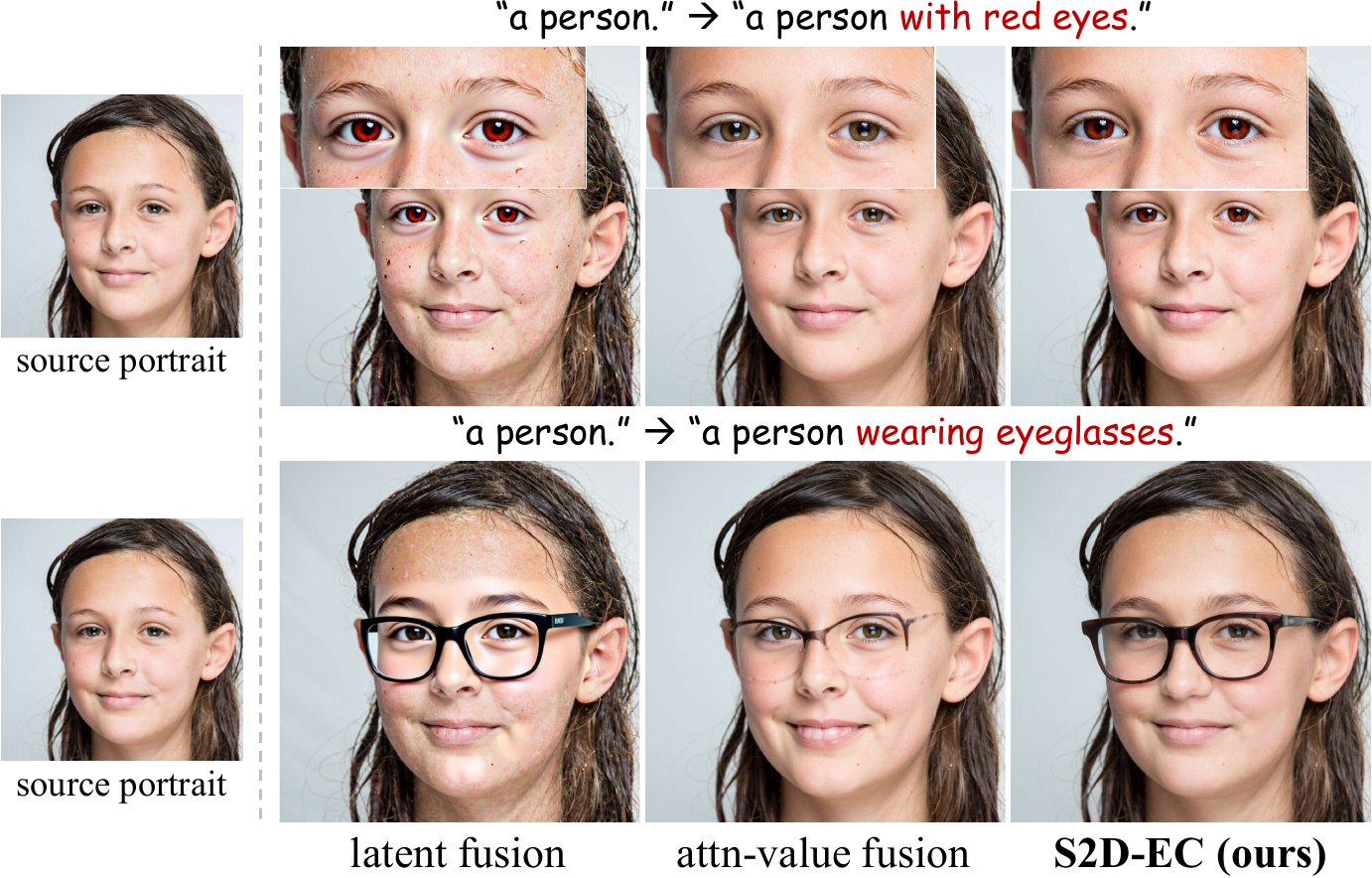}
    \caption{\textbf{Editing results under different mask guidance strategies.} We show the latent fusion, attention value fusion, and our S2D-EC's results in each column.}
    \label{fig:mask_strategies}
    \vspace{-0.2cm}
\end{figure}

\subsection{Structure-to-Detail Edit Control}
    In addition to precise editing localization, the second criterion for portrait editing is appropriate content modification. An ideal portrait edit should adhere to the target prompt while maintaining naturalness and consistency with the source portrait. Therefore, another challenge is how to effectively guide the editing process using the identified editing regions. Currently, there are two main mask guidance strategies: latent fusion~\cite{corneanu2024latentpaint,lugmayr2022repaint} and attention value fusion~\cite{cao2023masactrl,wang2024taming}. We implement both strategies and visualize them in Figure~\ref{fig:mask_strategies} (the left two columns). Upon observation, we find that the mask-guided latent fusion strategy produces the correct overall face layout and strong editing effects, but at the cost of losing naturalness and details. In contrast, the attention value fusion strategy effectively preserves the source texture and details, but its edit effects are weak.
    
    Meanwhile, previous work has shown that probabilistic flow-based models focus on low-frequency information in early denoising steps and high-frequency information in later steps~\cite{si2024freeu}. We observe a similar phenomenon in portrait editing, where the earlier steps roughly define the facial contour structure, while the later steps refine the facial details and textures. The difference in the model's focus on facial structure and details/textures between the early and later stages aligns closely with the differences observed in latent fusion and attention value fusion editing discussed earlier. Based on this observation, we propose a Structure-to-Detail Edit Control (S2D-EC) strategy that decomposes the denoising process into a facial structuring stage and a facial detailing stage.

\begin{table*}
    \centering
    \begin{adjustbox}{max width=0.95\linewidth}
    \begin{tabular}{p{3cm}<{\centering}|p{1.2cm}<{\centering}p{1.2cm}<{\centering}p{1.5cm}<{\centering}|p{1.2cm}<{\centering}p{1.2cm}<{\centering}p{1.2cm}<{\centering}p{1cm}<{\centering}p{2.2cm}<{\centering}}
        \toprule
        \multirow{2}*{Method} & \multicolumn{3}{c|}{Editing-related metrics} & \multicolumn{5}{c}{Preservation-related metrics} \\
        \cline{2-9}
         & $S_{\text{CLIP}}$ $\uparrow$ &$S^{dir}_{\text{CLIP}}$ $\uparrow$ & AttrEdit $\uparrow$ & PSNR $\uparrow$ & LPIPS $\downarrow$ & SSIM $\uparrow$ & ID $\uparrow$ & AttrPreserve $\uparrow$ \\
        \midrule
        StyleCLIP & 23.01 & 0.034 & 0.6504 & 18.29 & 0.1811 & 0.5459 & 61.26 & 0.9137 \\
        MasaCtrl & 21.41 & 0.054 & 0.6352& 16.85 & 0.2106 & 0.6455 & 32.40 & 0.8556 \\
        InstructPix2Pix & 22.00 & 0.073 & \textbf{0.7292}& 22.41 & 0.1153 & 0.7850 & 61.33 & 0.9065   \\
        MagicBrush & 22.63 & 0.080 & 0.7040& 25.72 & 0.0677 & 0.8459 & 72.79 & 0.9322 \\
        SmartEdit & 22.61 & 0.067 & 0.7260 & 21.47 & 0.1157 & 0.7141 & 57.64 & 0.9076 \\
        RF-Edit & 22.48 & 0.052 & 0.6576 & 23.50 & 0.1291 & 0.8051 & 49.15 & 0.8942   \\
        \cdashline{1-9}[1pt/1pt]
        Flux-Sculptor (ours)  & \textbf{23.04} & \textbf{0.082} & 0.7144& \textbf{28.56} & \textbf{0.0398} & \textbf{0.9238} & \textbf{74.11} & \textbf{0.9455} \\
        \bottomrule
    \end{tabular}
    \end{adjustbox}
    \caption{\textbf{Quantitative evaluations on CelebA-Edit.} We employ comprehensive evalation on both editing-related ($S_{\text{CLIP}}$: CLIP score, $S^{dir}_{\text{CLIP}}$: directional CLIP score, AttrEdit: attribute editing accuracy) and preservation-related (PSNR, LPIPS, SSIM, ID: face identity cosine similarity, AttrPreserve: attribute preservation accuracy) metrics.}
    \label{tab:comparison}
    \vspace{-0.3cm}
\end{table*}

    \noindent\textbf{Facial structuring stage.}
    During facial structuring stage, we adopt mask-guided latent fusion for spatial control. The editing region is denoted as $\mathcal{M}$, and the reconstruction region is $(1-\mathcal{M})$. Begining with the inverted latent $Z_{t_N}$, we inject the information from the reconstruction region of the source portrait into the editing flow. At timestep $t_{i}, i\in \{N,\cdots, N-T+1\}$, we first obtain the corresponding source latent $Z_{t_i}$ through linear interpolation between source latent $Z_{t_0}$ and the inverted latent $Z_{t_N}$:
    \begin{equation}
        Z_{t_{i-1}} = (1-t_{i-1})\cdot Z_{t_0} + t_{i-1}\cdot Z_{t_N}.
    \end{equation}
    We then follow RF-Solver~\cite{wang2024taming} to get the target latent $\tilde{Z}_{t_{i-1}}$:
    \begin{equation}
        \begin{aligned}
            \tilde{Z}_{t_{i-1}} = 
            \tilde{Z}_{t_{i}} & + (t_{i-1} - t_{i})v_{\theta}(\tilde{Z}_{t_{i}}, t_{i}) \\
            &+\frac{1}{2}(t_{i-1} - t_{i})^2 v^{(1)}_{\theta}(\tilde{Z}_{t_{i}}, t_{i}),
        \end{aligned}
    \end{equation}
    \begin{equation}
        \begin{aligned}
            v^{(1)}_{\theta}(\tilde{Z}_{t_{i}}, t_{i})
            &= \frac{v_{\theta}(\tilde{Z}_{t_{i}+\Delta t}, t_{i}+\Delta t) - v_{\theta}(\tilde{Z}_{t_{i}}, t_{i})}{\Delta t}.
        \end{aligned}
    \end{equation}
    Finally, we fuse the reconstruction region of source latent with the editing region of target latent to get the fused target latent $\tilde{Z}_{t_{i-1}}$ as the input of next DiT block:
    \begin{equation}
        \tilde{Z}'_{t_{i-1}} = \mathcal{M}\circ \tilde{Z}_{t_{i-1}} + (1- \mathcal{M})\circ Z_{t_{i-1}}.
    \end{equation}

    \noindent\textbf{Facial detailing stage.}
    After the $T$-step facial structuring stage, we adopt mask-guided attention value fusion in the detailing stage. Inspired by RF-Edit~\cite{wang2024taming}, we record the value features $\{V_{t_j,l}\}^{m}_{l=1}$ and $\{V_{t_j+\Delta t_i,l}\}^{m}_{l=1}$ from the self-attention layers in the last $m$ Transformer blocks at each timestep $t_j, j\in\{N-T, \dots 1\}$ during inversion. These value features capture the rich facial details and texture information of the source portrait, which is crucial for achieving a natural and identity-preserving edit.

    Then, we need to inject the source details from the reconstruction regions to the target attention values $\{\tilde{V}_{t_j,l}\}^{m}_{l=1}$. For the last $m$ Transformer blocks, the value features $V_{t_j,l}$ are the concatenation of the text value tokens $V^{\text{text}}_{t_j,l}$ and image value tokens $V^{\text{img}}_{t_j,l}$:
    \begin{equation}
        V_{t_j,l} = \left[ V^{\text{text}}_{t_j,l}, V^{\text{img}}_{t_j,l} \right],
    \end{equation}
    where $[\cdot]$ means token-wise concatenation. We then apply the mask-guided fusion between the source attention value and target attention value: 
    \begin{equation}
        \tilde{V'}^{\text{img}}_{t_j,l} = \mathcal{M}\circ \tilde{V}^{\text{img}}_{t_j,l} + (1- \mathcal{M})\circ V^{\text{img}}_{t_j,l}.
    \end{equation}
    Finally, we replace the image part of target attention value $\tilde{V}^{\text{img}}_{t_j,l}$ with the fused image attention value $\tilde{V'}^{\text{img}}_{t_j,l}$, while retaining its text part $\tilde{V}^{\text{text}}_{t_j,l}$ to obtain the fused attention values $\tilde{V}'_{t_j,l}$, which will be input to the next Transformer block:
    \begin{equation}
        \tilde{V}'_{t_j,l} = \left[ \tilde{V}^{\text{text}}_{t_j,l}, \tilde{V'}^{\text{img}}_{t_j,l} \right].
    \end{equation}
    
    In conclusion, S2D-EC employs a two-stage structure-to-detail spatial control that aligns with the denoising process of probabilistic flow-based models. As visualized in the last column of Figure~\ref{fig:mask_strategies}, our S2D-EC strategy effectively achieves appropriate content modification with satisfactory strength and naturalness.

\section{Experiments}
\subsection{Experimental Setup}
    \noindent\textbf{Dataset.}
    \label{sec:text2mask_dataset}
    We establish a Text2Mask dataset to train our Prompt-Aligned Spatial Locator, using 80,000 portrait images from FFHQ1024~\cite{karras2019style}. For each image, we first define 18 fine-grained face regions (e.g., cheeks, eyes, nose, lips, etc.) and extract their corresponding masks using face parsing and landmark detection techniques. Then, for each face region, we manually attach a target text randomly sampled from our pre-generated text list related to the region (e.g., "a person with black hair" for the "hair" region). As a result, we obtain 1.25 million portrait samples with corresponding target texts and masks for training PASL. The Text2Mask dataset is described in supplementary materials A.1.

    \noindent\textbf{Implementation details.}
    Flux-Sculptor adopts FLUX.1-dev~\cite{flux2024} as the foundation model, with its weights frozen. The total number of inversion and editing steps is set to $N=30$. The PASL consists of a CNN-based image encoder, the frozen text encoder of AlphaCLIP~\cite{sun2024alpha}, and a multi-modal projector. The image encoder and projector are trained for 32 epochs from scratch on a single H800-80G GPUs, with an input resolution to be $512\times 512$. We use the AdamW~\cite{loshchilov2017fixing} optimizer with a weight decay of $2\times10^{-2}$. The learning rate is set to $1\times 10^{-4}$, with a cosine learning rate scheduler. The PASL network is detailed in the Supplementary Materials A.2.

    \noindent\textbf{The CelebA-Edit benchmark.}
    We utilize CelebA-HQ~\cite{karras2017progressive} and its facial attribute labels to establish our portrait editing benchmark named CelebA-Edit. We select 24 clearly defined facial attributes and generate corresponding source and target prompts. For each attribute, we sample 100 portraits from CelebA-HQ, resulting in a total of 2,400 testing samples for evaluation. 
    
    \noindent\textbf{Evaluation metrics.}
    Aligning with our previous discussion, we design both editing-related and preservation-related metrics to ensure a comprehensive evaluation of portrait editing. For editing, we use the CLIP score~\cite{radford2021learning} and directional CLIP score~\cite{kim2022diffusionclip} to measure the coarse-grained text-edit alignment. Additionally, we employ attribute editing accuracy~\cite{wang2024maniclip} to measure fine-grained alignment. For preservation, we use PSNR~\cite{huynh2008scope}, LPIPS~\cite{zhang2018unreasonable}, SSIM~\cite{wang2004image} and face identity similarity~\cite{deng2019arcface} to assess overall consistency, and attribute preservation accuracy to ensure the retention of facial attributes and details. A pretrained facial attribute classifier~\cite{he2017adaptively} is used to obtain the attribute editing and preservation accuracies.

\begin{figure*}
    \centering
    \includegraphics[width=0.95\linewidth]{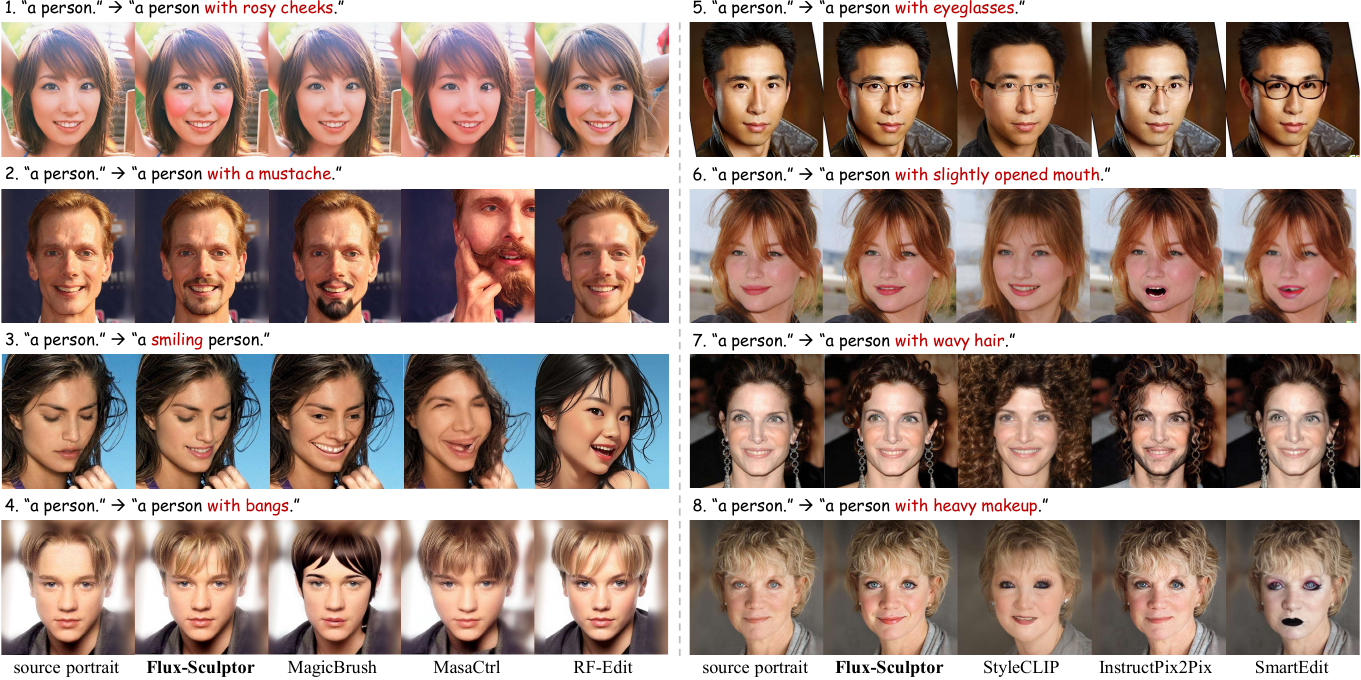}
    \caption{\textbf{Qualitative comparisons.} We conduct portrait editing with eight different text prompts and visualize the results of our Flux-Sculptor and all competitors (left column: MagicBrush, MasaCtrl, RF-Edit; right column: StyleCLIP, InstructPix2Pix, SmartEdit).}
    \label{fig:visualization}
    \vspace{-0.2cm}
\end{figure*}

\subsection{Comparison with State-of-the-Art Methods}
\label{sec:comparison}
    \noindent\textbf{Objective evaluation.}
    We compare Flux-Sculptor with several state-of-the-art text-based image editing methods, including StyleCLIP~\cite{patashnik2021styleclip}, MasaCtrl~\cite{cao2023masactrl}, InstructPix2Pix~\cite{brooks2023instructpix2pix}, MagicBrush~\cite{zhang2023magicbrush}, SmartEdit~\cite{huang2024smartedit}, and RF-Edit~\cite{wang2024taming}. The quantitative results are shown in Table~\ref{tab:comparison}. By observing the editing-related metrics, Flux-Sculptor yields the best $S_{\text{CLIP}}$, $S^{dir}_{\text{CLIP}}$, and competitive attribute editing (AttrEdit) performance among all methods, demonstrating its strong editing capability. It is worth noting that diffusion-based methods, such as InstructPix2Pix and SmartEdit, get high AttrEdit results at the cost of low faicial reconstruction performances, where the loss outweighs the gain. Furthermore, according to the preservation-related metrics, Flux-Sculptor outperforms all other methods across five metrics by a significant margin. This highlights that Flux-Sculptor effectively maintains facial details and identity information while accommodating various editing requirements. More information about AttrEdit and AttrPreserve is in Supplementary Materials B. The quantitative comparison statistically proves that our method strikes a good balance between reconstruction fidelity and editing flexibility.

\begin{figure}[t]
    \centering
    \includegraphics[width=0.95\linewidth]{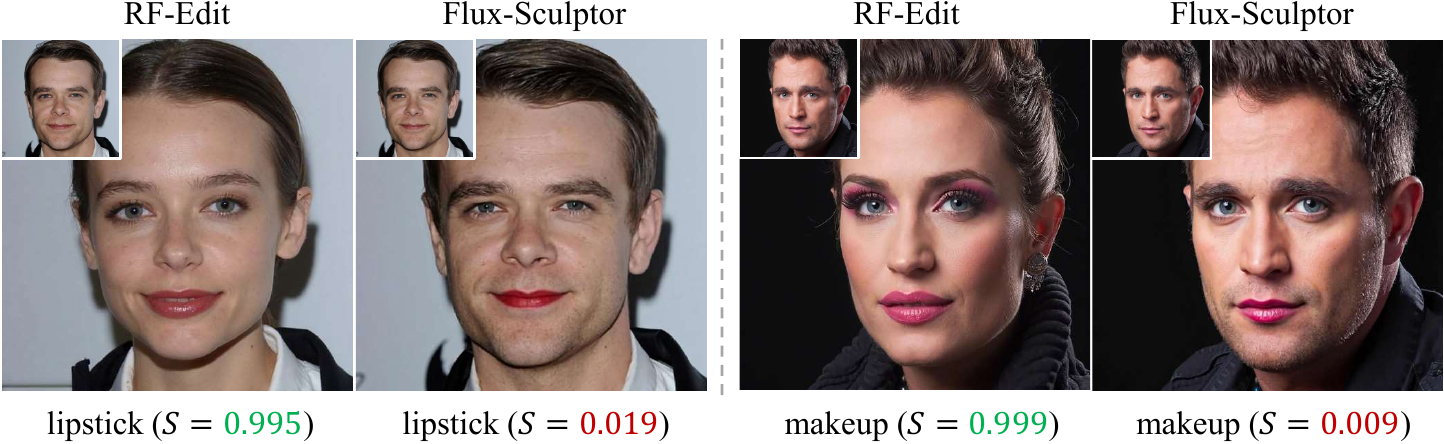}
    \caption{\textbf{Gender-biased attribute editing.} We edit male portraits with ``wearing lipstick'' and ``with makeup'' on RF-Edit and Flux-Sculptor, and visualize classification scores $S$ for comparison.}
    \label{fig:gender_debias_vis}
    \vspace{-0.3cm}
\end{figure}

\begin{figure*}
    \centering
    \includegraphics[width=0.9\linewidth]{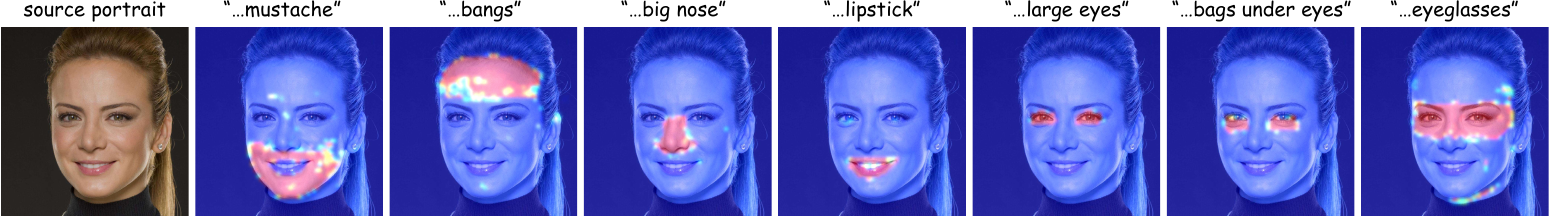}
    \caption{\textbf{Text-driven editing region localization of PASL.} Given the source portrait and seven different text prompts, our PASL can provide fine-grained and accurate spatial responses to various user prompts.}
    \label{fig:text2mask}
    \vspace{-0.3cm}
\end{figure*}

    \noindent\textbf{Subjective evaluation on gender-biased attributes.}
    When evaluating attribute editing (AttrEdit), we found that current attribute classifiers struggle with gender-biased attributes. Additionally, some image editing methods fail to decouple these biased attributes from gender effectively. As shown in Figure~\ref{fig:gender_debias_vis}, during gender-biased attribute editing, RF-Edit will transform a male portrait into a female one, achieving a high classification score. In contrast, Flux-Sculptor can successfully apply lipstick and makeup to a male portrait without altering its gender, but result in a very low classification score. Therefore, in the objective evaluation, we only sampled female portraits for editing ``big lips'', ``lipstick'' and ``makeup'' to ensure the correctness of automatic evaluation. Regarding those three gender-biased attributes, we sample 30 portraits of different genders for editing each, amounting to a total of 90 images. We then recruited 20 participants to conduct a subjective human evaluation, with statistical results from the 1,800 choices displayed in Table~\ref{tab:gender_biased_sub}. Both human evaluation and visual results validate that Flux-Sculptor effectively decouples attribute editing from the gender, demonstrating a more flexible editing ability.

\begin{table}[t]
    \centering
    \begin{adjustbox}{max width=0.95\linewidth}
    \begin{tabular}{p{2.9cm}<{\centering}|p{1.2cm}<{\centering}p{1.3cm}<{\centering}p{1.4cm}<{\centering}}
        \toprule
        \multirow{2}*{Method} & \multicolumn{3}{c}{Human evaluation} \\
        \cline{2-4}
         & big lip $\uparrow$ & lipstick $\uparrow$ & makeup $\uparrow$ \\
        \midrule
        InstructPix2Pix         & 8.00\% & 3.33\% & 7.67\%  \\
        MagicBrush              & 2.33\% & 22.67\% & 8.67\%  \\
        RF-Edit                 & 15.67\% & 6.00\% & 8.00\%  \\
        \cdashline{1-4}[1pt/1pt]
        Flux-Sculptor (ours)    & \textbf{74.00\%} & \textbf{68.00\%} & \textbf{75.67\%}  \\
        \bottomrule
    \end{tabular}
    \end{adjustbox}
    \caption{\textbf{Subjective evaluation results.} We illustrate the percentage of participants who chose each method for every attribute.}
    \label{tab:gender_biased_sub}
    \vspace{-0.3cm}
\end{table}

    \noindent\textbf{Qualitative comparison.}
    We present eight qualitative comparisons in Figure~\ref{fig:visualization} to further validate the effectiveness of our method. In examples 1 and 8, while other competitors either introduce excessive appearance alteration or fail to achieve satisfactory editing effects, Flux-Sculptor successfully applies natural and visually appealing modifications to facial details (e.g., “rosy cheeks” and “makeup”). Additionally, our method demonstrates a significant advantage in facial structure editing, including expression (examples 2 and 6) and decoration (example 5) modification. Existing diffusion-based methods (e.g., MasaCtrl, InstructPix2Pix, and MagicBrush)  often produce distorted facial features due to their ineffective masking strategies, while RF-Edit and the GAN-based StyleCLIP struggle to preserve fine facial details. Our method effectively fulfills the user’s editing intent while maintaining consistency and realism in the edited images. Furthermore, examples 2, 4, and 7 highlight our model’s capability to accurately edit complex attributes such as hair and beards. In summary, by leveraging precise editing localization and structure-to-detail edit control, Flux-Sculptor achieves high-quality portrait editing while faithfully preserving facial identity and details.

\begin{figure}[t]
    \centering
    \includegraphics[width=0.9\linewidth]{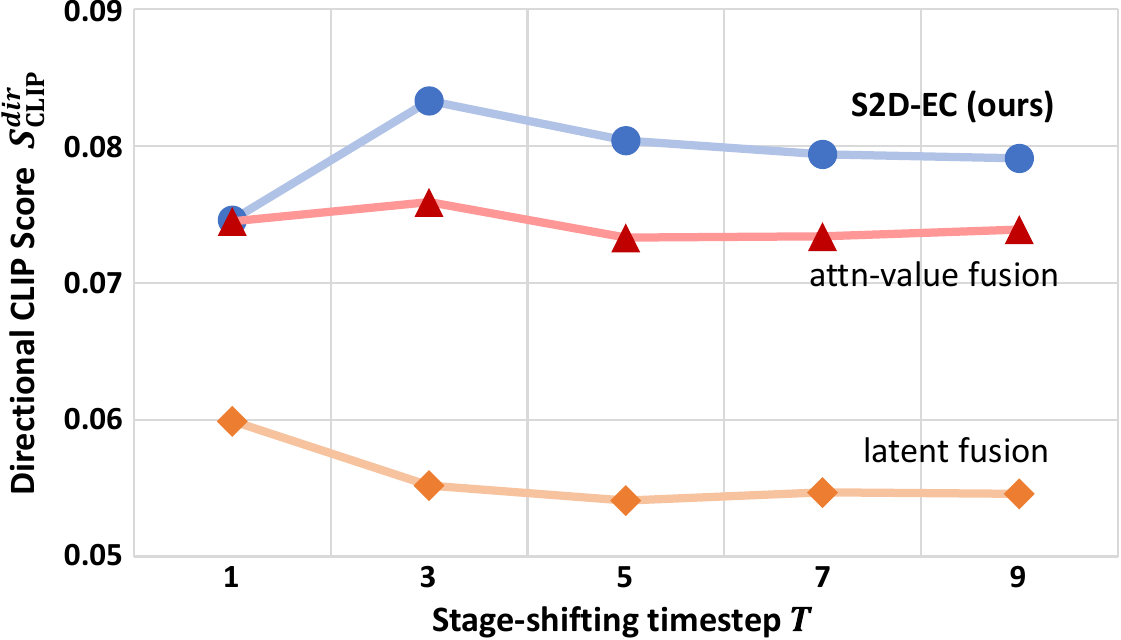}
    \caption{\textbf{Ablation study on stage-shifting timestep $T$.} We report $S^{dir}_{\text{CLIP}}$ of the latent fusion, attention value fusion and our S2D-EC strategies on $T=1,3,5,7,9$.}
    \label{fig:ablation_T}
    \vspace{-0.3cm}
\end{figure}

\subsection{Ablation Study}
    \noindent\textbf{$T$ for different denoising control strategies.}
    The S2D-EC strategy has the hyperparameter $T$ to control the switching timesteps between the facial structuring stage and the detailing stage. In this section, we experiment with how $T$ affects different denoising control strategies. Different denoising settings are as follows. 1) \textit{latent fusion} adopts mask-guided latent fusion in the former $T$ denoising steps, similar to methods~\cite{corneanu2024latentpaint,lugmayr2022repaint,couairon2022diffedit}; 2) \textit{attn-value fusion} adopts mask-guided attention value fusion in the latter $T$ denoising steps, similar to methods~\cite{cao2023masactrl,wang2024taming}; 3) \textit{our S2D-EC}. We illustrate their $S_{\text{CLIP}}$ on our validation set under different values of $T$. The performances of S2D-EC and attention-value fusion initially rise and then fall, peaking at $T=3$, while latent fusion's performance declines as $T$ increases. Overall, S2D-EC provides the most effective spatial control among all strategies, and an appropriate value of $T$ can positively affect the edits.
    
    \noindent\textbf{Proposed module and strategy.}
    We further conduct an ablation study on both the PASL module and the S2D-EC stretegy. We use $S_{\text{CLIP}}$ and Face ID to reflect the editing and preservation performances, respectively. When PASL is not activated, fusion is applied to the entire feature maps. It can be observed that S2D-EC's structure-to-detail portrait editing achieves a good balance between editing and preservation. Subsequently, by further spatially locating the editing regions, the preservation abilities of all fusion strategies are improved by a large margin (8.13\%, 19.87\%, and 5.08\%, respectively). Meanwhile, the editing performance gains slight improvements, as a better ability to preserve facial details benefits the generation of edited images that closely adhere to the detailed target prompts.

\begin{table}[t]
    \centering
    \begin{adjustbox}{max width=0.9\linewidth}
    \begin{tabular}{p{1cm}<{\centering}|p{2.5cm}<{\centering}|p{1.5cm}<{\centering}|p{1.5cm}<{\centering}}
        \toprule
        \textbf{PSAL} & \textbf{Fusion Strategy} & $S_{\text{CLIP}} \uparrow$ & ID $\uparrow$ \\
        \midrule
                    & /                 & 20.30 &  18.02 \\
                    & latent fusion     & 21.40 &  24.61 \\
                    & attn-value fusion & 22.82 &  49.15 \\
                    & S2D-EC (ours)     & 23.01 &  69.03 \\
        \cdashline{1-4}[1pt/1pt]
        \checkmark  & latent fusion     & 21.47 & 32.74  \\
        \checkmark  & attn-value fusion & 22.93 & 69.02 \\
        \checkmark  & S2D-EC (ours)     & \textbf{23.04} & \textbf{74.11} \\
        \bottomrule
    \end{tabular}
    \end{adjustbox}
    \caption{\textbf{Ablation study on the proposed modules.} We show the CLIP score and Face ID performances on CelebA-Edit.}
    \label{tab:ablation}
    \vspace{-0.3cm}
\end{table}

\subsection{Text-Driven Editing Localization}
    To further validate the precise editing localization ability of Flux-Sculptor, we visualize the text-driven editing region localization results of our PASL module in Figure~\ref{fig:text2mask}. While text-driven facial localization remains a challenging task for current general segmentation methods~\cite{kirillov2023segany,ravi2024sam2}, our PASL can provide fine-grained and accurate spatial responses to the users' text prompts, paving the way for our ideal editing. More insightful Experiments about localization and editing are available in the Supplementary Materials A.3-4 and C.

\section{Conclusion}
    We propose Flux-Sculptor for text-driven portrait editing. To achieve precise editing localization, we construct PASL to precisely locate text-related regions. To ensure appropriate content modification, we design the S2D-EC strategy that spatially controls the editing process through mask-guided fusion on the latent and attention value. Extensive experiments validate its superior performance for text-driven, rich-attribute portrait editing. We believe that Flux-Sculptor holds great potential for practical applications and can provide valuable insights for the future research.

\newpage
{
    \small
    \bibliographystyle{ieeenat_fullname}
    \bibliography{main}
}

\clearpage
\appendix

\begin{strip}
  \centering
  \Large \textbf{Flux-Sculptor: Text-Driven Rich-Attribute Portrait Editing through Decomposed Spatial Flow Control} \\
  \vspace{2pt}
  Supplementary Materials
  \vspace{1em}

\end{strip}

\subfile{supple}

\end{document}

%% file: preamble.tex
\usepackage{adjustbox}
\usepackage{algorithm}
\usepackage{subcaption}
\usepackage{algorithmic}
\usepackage{amsmath}
\usepackage{amssymb}
\usepackage{arydshln}
\usepackage{array}
\usepackage{caption}
\usepackage{color}
\usepackage{enumitem}
\usepackage{epsfig}
\usepackage{epstopdf}
\usepackage{flushend}
\usepackage{graphicx}
\usepackage{makecell}
\usepackage{multirow}
\usepackage{multicol}
\definecolor{ngreen}{RGB}{17, 173, 30}
\usepackage{soul}
\usepackage{tabularx}
\usepackage{times}
\usepackage{utfsym}
\usepackage[normalem]{ulem}
\usepackage{xcolor}
\usepackage{bm}

%% file: supple.tex


\section{Additional Details and Experiments about Text-Driven Editing Region Localization}
    Text-driven editing localization is an important part of our framework. In this section, we will provide additional informative details and insightful experiments, including the dataset establishment, the network design, and experimental results of localization and editing.

\begin{figure}[H]
    \centering
    \includegraphics[width=0.95\linewidth]{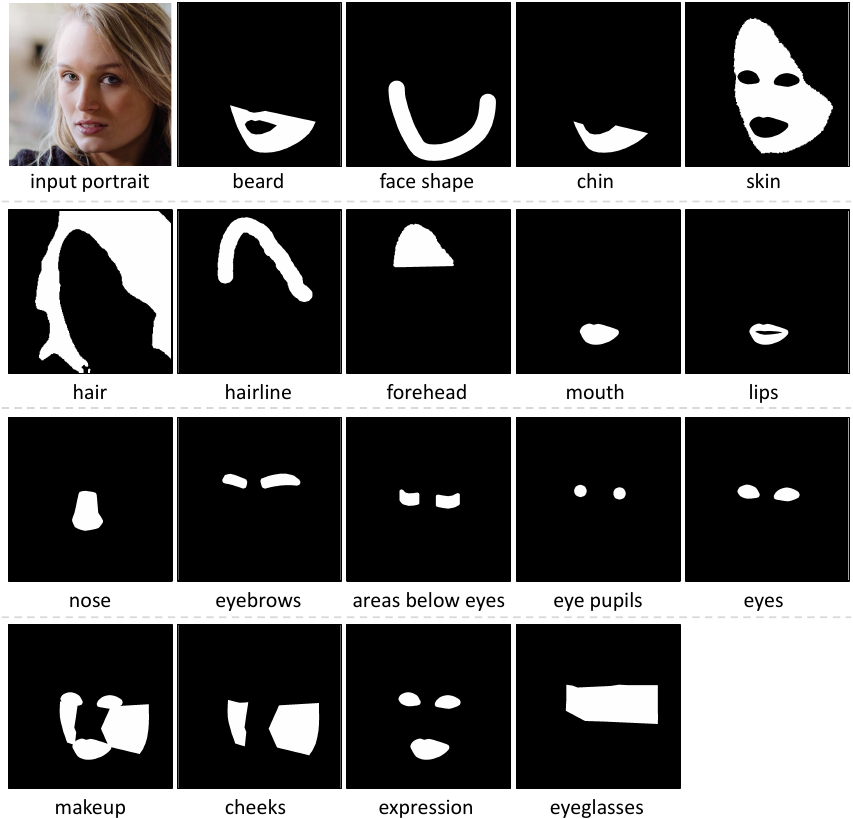}
    \caption{\textbf{Pre-defined and extracted face editing regions.}}
    \label{fig:dataset_mask}
\end{figure}
\vspace{-3pt}

\begin{figure}[t]
    \centering
    \includegraphics[width=0.9\linewidth]{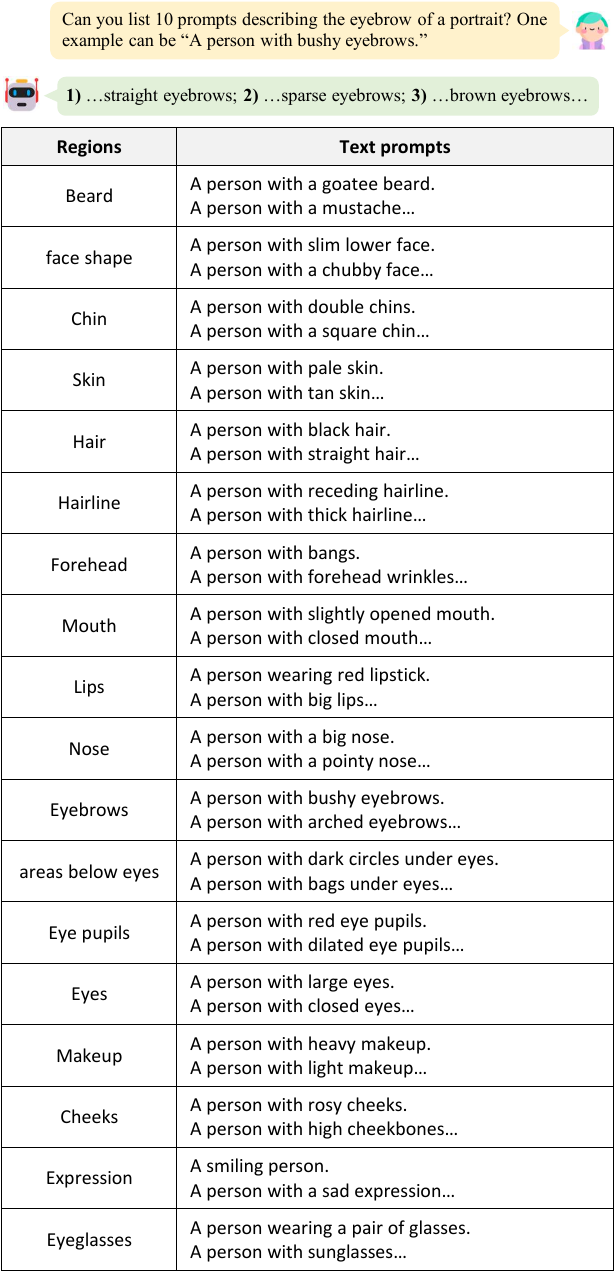}
    \caption{\textbf{Generated text prompts for each region.}}
    \label{fig:text_generate}
\end{figure}

\begin{figure*}
    \centering
    \includegraphics[width=0.95\linewidth]{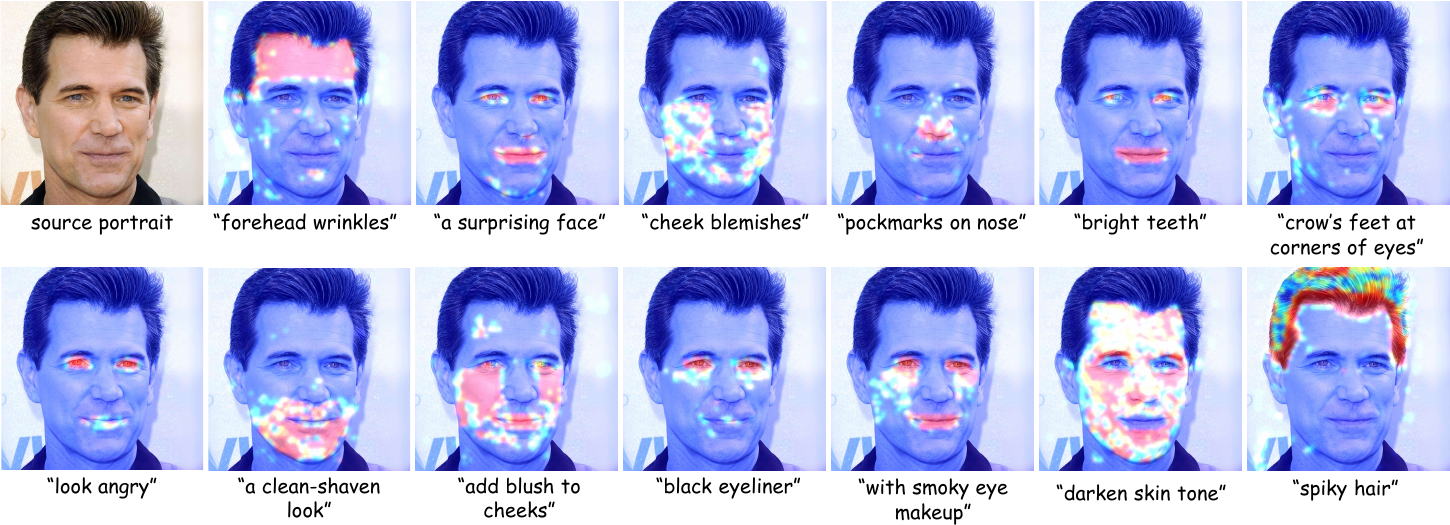}
    \caption{\textbf{Editing region localization results according to open-set text prompts.}}
    \label{fig:ood_text2mask}
\end{figure*}

\vspace{-3pt}
\subsection{Text2Mask Dataset Establishment}
    To achieve the challenging fine-grained facial localization, we establish a Text2Mask dataset to train our Prompt-Aligned Spatial Locator. We use 80,000 high-quality portrait images from FFHQ1024 as the source portraits. Then, for each portrait, we adopt a face landmark detector and face parser tools to extract a total of 18 representative face regions for editing, as illustrated in Figure~\ref{fig:dataset_mask}. Upon the pre-defined mask, we need to construct text-mask pairs further for training. During construction, we first manually write the target prompts according to the facial attributes labels of CelebA-HQ. Then, we use these manual prompts as few-shot examples to guide ChatGPT to generate rich region-related target prompts, as is shown on top of Figure~\ref{fig:text_generate}. We generate ten various text prompts related to the target regions, and list two for each region in the table in Figure~\ref{fig:text_generate}. Based on the prepared mask and prompts, we construct text-mask pairs by attaching a random text to each mask. Consequently, we obtain a total of 1.35 million samples after filtering bad cases to establish our Text2Mask dataset.

\subsection{Network Structure of the PASL}
    The PASL module consists of a CNN-based image encoder, a pretrained text encoder from AlphaCLIP, and a multi-modal projector. We will describe them in detail.
\begin{table}[t]
    \centering
    \begin{adjustbox}{max width=\linewidth}
    \begin{tabular}{cccc}
        \toprule
        \textbf{Layer} & \textbf{In $\bm{\to}$ Out Dim} & \textbf{Stride} & \textbf{Activation}   \\
        \midrule
        conv-1       & $3\to 64$     & 1 & LeakyReLU    \\
        conv-2       & $64\to 128$   & 2 & LeakyReLU    \\
        conv-3       & $128\to 256$  & 2 & LeakyReLU    \\
        conv-4       & $256\to 512$  & 2 & LeakyReLU    \\
        conv-5       & $512\to 512$  & 1 & LeakyReLU    \\
        conv-6       & $512\to 512$  & 1 & LeakyReLU    \\
        \bottomrule
    \end{tabular}
    \end{adjustbox}
    \caption{\textbf{The network structure of the PASL's image encoder.} All layers are 2D convolution with padding=1, kernel size=$3\times 3$.}
    \label{tab:image_encoder}
\end{table}

    \noindent\textbf{Image encoder.}
    The goal of the image encoder is to efficiently extract fine-grained and discriminative facial features, we design its structure based on the convolutional neural networks, taking advantage of its data-efficient spatial local prior and concise structure. The network structure is shown in Table~\ref{tab:image_encoder}. The input resolution for the image encoder is set to $512\times 512$, and the output shape is $64\times 64\times 512$. 

\begin{table}[t]
    \centering
    \begin{adjustbox}{max width=0.8\linewidth}
    \begin{tabular}{ccc}
        \toprule
        \textbf{Layer}   & \textbf{In $\bm{\to}$ Out Dim} & \textbf{Activation}   \\
        \midrule
        linear-1 & $768\to 1024$   & ReLU    \\
        linear-2 & $1024\to 512$   & /    \\
        \bottomrule
    \end{tabular}
    \end{adjustbox}
    \caption{\textbf{The network structure of the PASL's multi-modal projector.} We use a multilayer perceptron for cross-modal projection.}
    \label{tab:mm_projector}
\end{table}

    \noindent\textbf{Multi-modal projector.}
    The multi-modal projector aims to align the text embeddings to the visual feature space. Its network structure is demonstrated in Table~\ref{tab:mm_projector}. The original output shape of AlphaCLIP is $1\times 768$. Multi-modal projector transforms it into $1\times 512$, which aligns with $f^{\text{img}}$'s dimension size. Finally, by expanding $f^{\text{text}}\in R^{512}$ to $\hat{f}^{\text{text}}\in R^{64\times 64\times 512}$ through broadcasting, we can compute the pixel-wise similarity between visual and language, yielding the final editing region mask $\mathcal{M}$. 

    \noindent\textbf{Overhead Analysis.}
    As described above, the PASL is designed to be quite lightweight. For validation, we conduct an overhead analysis on the submodule of PASL on a single H800-80G GPU. The results shown in Table~\ref{tab:overhead} show that our editing region localization process is very efficient and provides negligible overhead.
    
\begin{table}[t]
    \centering
    \begin{adjustbox}{max width=\linewidth}
    \begin{tabular}{p{4cm}<{\centering}|p{1cm}<{\centering}p{1cm}<{\centering}p{1cm}<{\centering}}
        \toprule
        
        \multirow{2}*{\textbf{Sub-Module}} & \textbf{FLOPS} & \textbf{Params} & \textbf{Latency}   \\
         & \textbf{(G)} & \textbf{(M)} & \textbf{(ms)}   \\
        \midrule
        Image Encoder           & 68.694  & 6.27     & 37.78    \\
        AlphaCLIP Text Encoder  & 4.360   & 56.67    & 25.90    \\
        Multi-Modal Projector   & 0.003   & 1.31     & 0.07    \\
        \cdashline{1-4}[1pt/1pt]
        Complete PASL           & 73.057  & 64.25    & 63.75    \\
        \bottomrule
    \end{tabular}
    \end{adjustbox}
    \caption{\textbf{Overhead Analysis on PASL.}}
    \label{tab:overhead}
\end{table}

\subsection{Open-Set Text to Mask Generation}
    In the mainscript, we show the generated mask for some common and in-distribution text prompts. In this section, we further visualize PASL's results on text prompts that the model hasn't seen in the training set. The results are vividly shown in Figure~\ref{fig:ood_text2mask}. By observation, PASL can still give appropriate spatial responses to these challenging out-of-distribution text prompts, proving its potential for broader application. This strong generalization ability mainly comes from AlphaCLIP's text encoder. We freeze its pretrained weights to preserve the text comprehension and spatial perception capabilities.
    
\begin{figure}[b]
    \centering
    \includegraphics[width=0.95\linewidth]{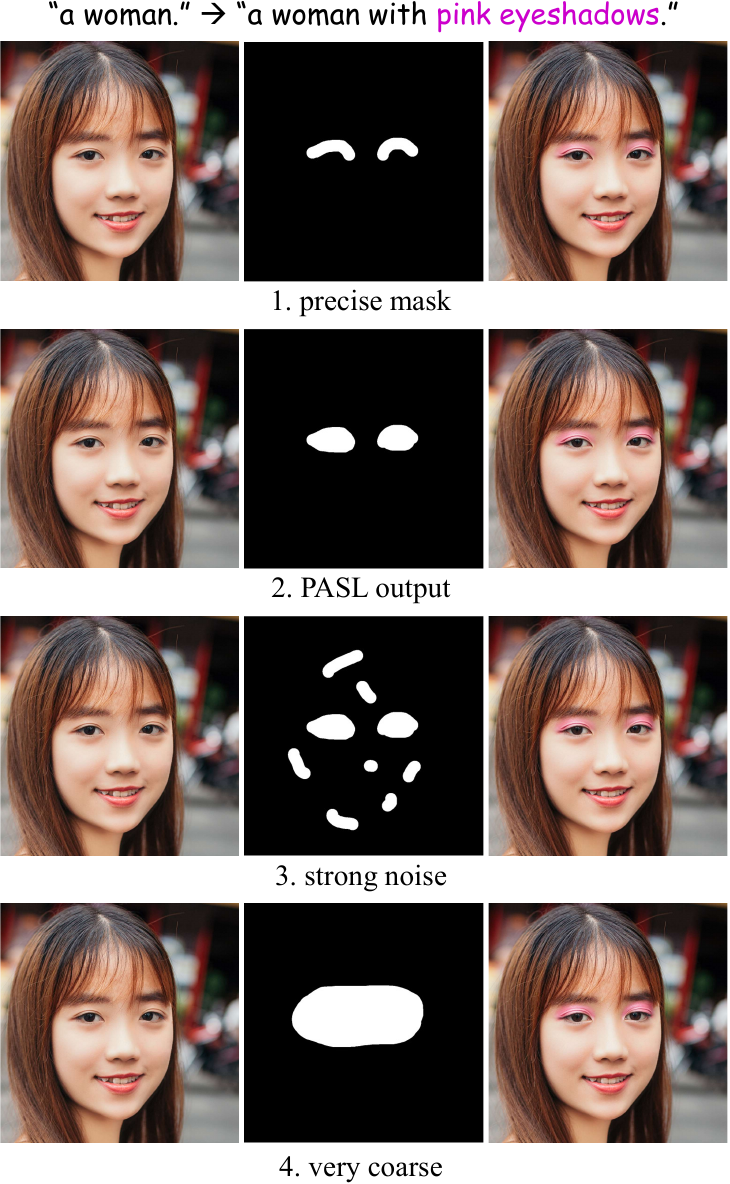}
    \caption{\textbf{Edits over masks with different noises.}}
    \label{fig:robust_mask_edit}
\end{figure}

\subsection{Robustness to Mask Guidance during Editing}
    In addition to PASL's accurate editing localization, our S2D-EC denoising strategy exhibits strong robustness to different variations of the mask's guidance which may happen in real scenarios. We show the editing results over masks with different kinds and degrees of noise in Figure~\ref{fig:robust_mask_edit}. Case 1 shows the perfect editing results with the most precise manual editing localization. In case 2, although PASL's output cannot be as accurate as the human-annotated mask, our S2D-EC strategy can still yield nearly perfect edits. In case 3, We add strong noises and scatters to the mask, and our method can ignore the irrelevant noisy locations and provide satisfying editing results. Finally, in case 4, we provide a very coarse mask with a large proportion of unrelated areas. Our method still generates an acceptable edit with a slight change to the looking direction. In conclusion, our structure-to-detail editing control is robust to different noisy masks, making it more applicable.

\begin{figure*}
    \centering
    \includegraphics[width=\linewidth]{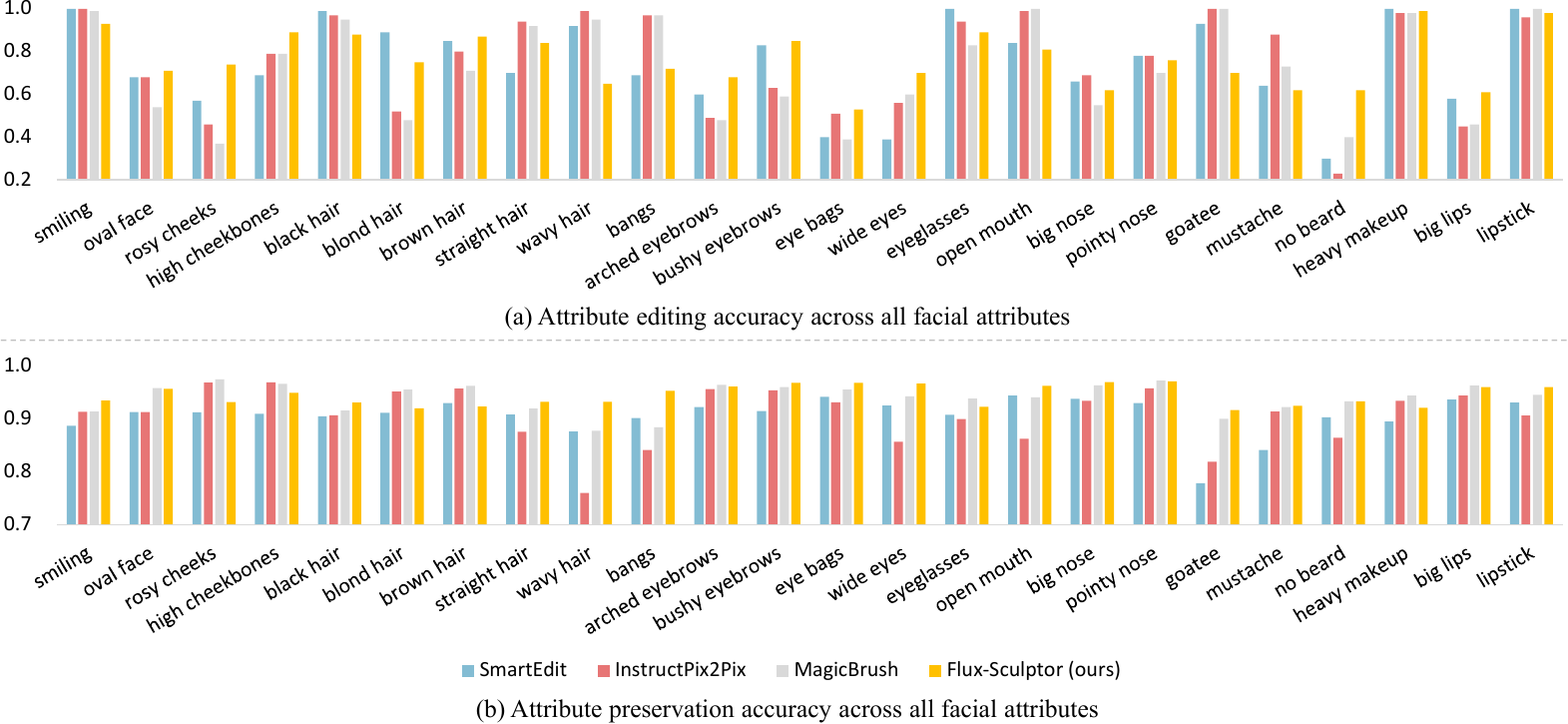}
    \caption{\textbf{Attribute-wise AttrEdit and AttrPreserve accuracy.} We exhibit the fine-grained editing and preservation performances of SmartEdit, InstructPix2Pix, MagicBrush, and our Flux-Sculptor across all 25 facial attributes.}
    \label{fig:attr_wise_acc}
\end{figure*}

\section{Fine-Grained Portrait Editing Metrics}
    In the manuscript's Section 4.2, we design both coarse-grained and fine-grained metrics to evaluate the editing and preservation abilities. These coarse-grained metrics have been widely adopted in image editing tasks. The fine-grained attribute-level metrics, including Attribute Editing Accuracy (AttrEdit) and Attribute Preservation Accuracy (AttrPreserve), are newly designed in this paper, with reference to previous portrait editing works. We adopt a facial attribute classifier trained on CelebA. The classifier achieves a mean accuracy of 91.80\% on the multi-attribute classification task, which can reflect whether the target attribute is edited or preserved.

\subsection{Metric Definition}
    \noindent\textbf{Attribute editing accuracy.}
    In CelebA-Edit, we choose 24 facial attributes from CelebA's annotations for editing. A good edited portrait should get a high classification score on the target attribute aligned with the text prompt. The AttrEdit metric is calculated as follows:
    \begin{equation}
        \text{AttrEdit}=\frac{1}{N}\sum_{i=1}^{N} \mathbb{I}(\tilde{s}^{\text{tgt}}_{i}>\tau),
    \end{equation}
    where $N$ is the dataset size and $i$ means the $i_{\text{th}}$ sample. $\tilde{s}^{\text{tgt}}_{i}$ is the edited portrait's classification score on the target attribute, with classification threshold $\tau=0.1$. AttrEdit ranges from 0 to 1 and higher values indicate better performance.

    \noindent\textbf{Attribute preservation accuracy.}
    To evaluate detail preservation ability, all irrelevant facial attributes in the edited image should be the same as those in the source image. Therefore, $\text{AttrPreserve}$ is calculated as below:
    \begin{equation}
        \begin{aligned}
            \text{AttrPreserve} &= \frac{\sum_{j=1}^{M_i} \mathbb{I}\left[\mathbb{I}(s_{i,j}>\tau')=a^{\text{pres}}_{i,j}\right]}{\sum_{i=1}^{N} M_{i}}, \\
        \end{aligned}
    \end{equation}
    where $M_i$ is the number of preservation attributes of the $t_{\text{th}}$ sample, $s_{i,j}$ is the classification score of the $j_{\text{th}}$ preseration attribute of the $i_{\text{th}}$ sample, and $a^{\text{pres}}_{i,j}$ is its corresponding ground truth source portrait label. AttrEdit also ranges from 0 to 1, with a higher value indicating better performance.

\subsection{Attribute-Wise Fine-Grained Performance}
    We calculate the attribute editing and preservation accuracies of SmartEdit, InstructPix2Pix, MagicBrush and our Flux-Sculptor. Figure~\ref{fig:attr_wise_acc}(a) shows the AttrEdit results. We show the four best methods. By observation, the accuracies on these coarse attributes with large modification areas (e.g., hairstyle, makeup) are consistently higher than those fine-grained local modifications (e.g., big lips, eye). Although our Flux-Sculptor doesn't achieve the best editing performance, we can find that it can achieve outstanding editing effects on many challenging attributes where other methods get low scores, such as ``oval face'', ``rosy cheeks'', ``eyebrows'', and ``eye bags''. Through visualization, Flux-Sculptor tends to give natural and moderate edits, while the editing effects of other diffusion-based methods are usually more pronounced and even exaggerated. This may account for their higher editing accuracies. We then show the AttrPreserve in Figure~\ref{fig:attr_wise_acc}(b). It can be observed that our method consistently outperforms the other competitors, validating its strong detail and identity preservation capability on the attribute level.

\section{Extended Functionalities of Flux-Sculptor}
    Besides the text-driven portrait editing task discussed in our main script, Flux-Sculptor also shows great potential for more practical applications. In this section, we will introduce two extended functionalities, which are \textit{mask-guided portrait editing} and \textit{multi-attribute portrait editing}.

\subsection{Mask-Guided Text-Driven Portrait Editing}
\begin{figure}[t]
    \centering
    \includegraphics[width=0.95\linewidth]{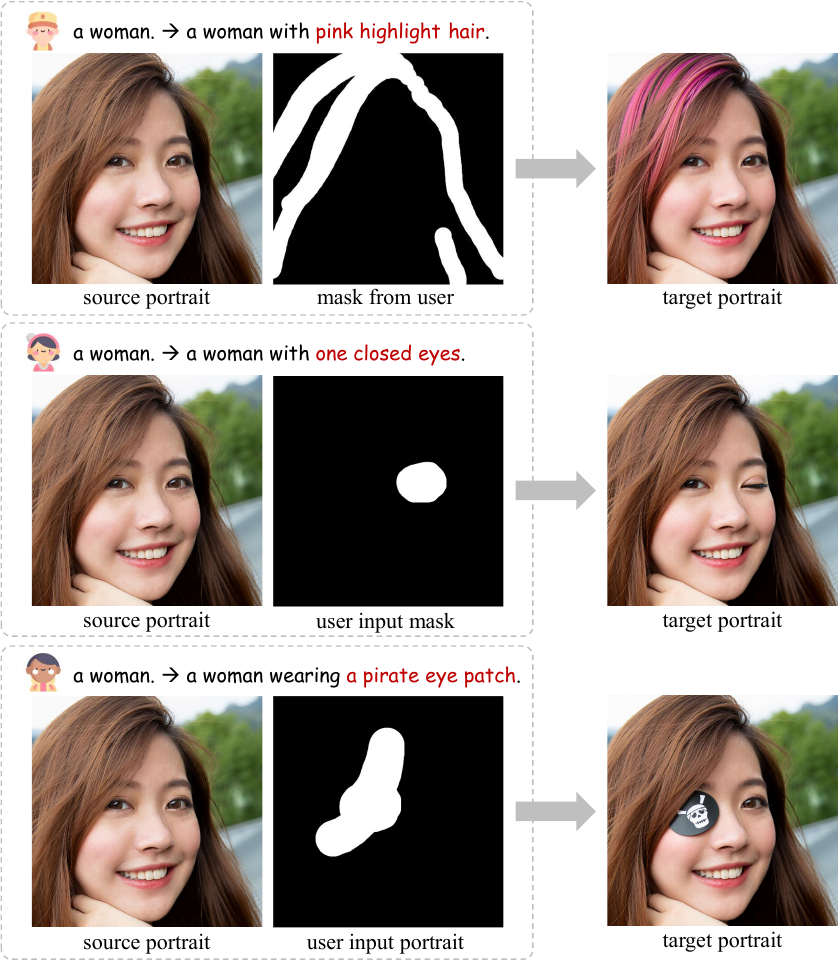}
    \caption{\textbf{Customized editing based on user input mask.}}
    \label{fig:mask_guided_user}
\end{figure}

    During text-driven portrait editing, the editing mask is automatically generated by our PASL. Meanwhile, our method can also accept the user's input mask as spatial guidance. The manual mask can be more personalized and precise, satisfying special editing requirements, as illustrated in Figure~\ref{fig:mask_guided_user}.
    
\subsection{Multi-Attribute Text-Driven Portrait Editing}
\begin{figure}[t]
    \centering
    \includegraphics[width=0.95\linewidth]{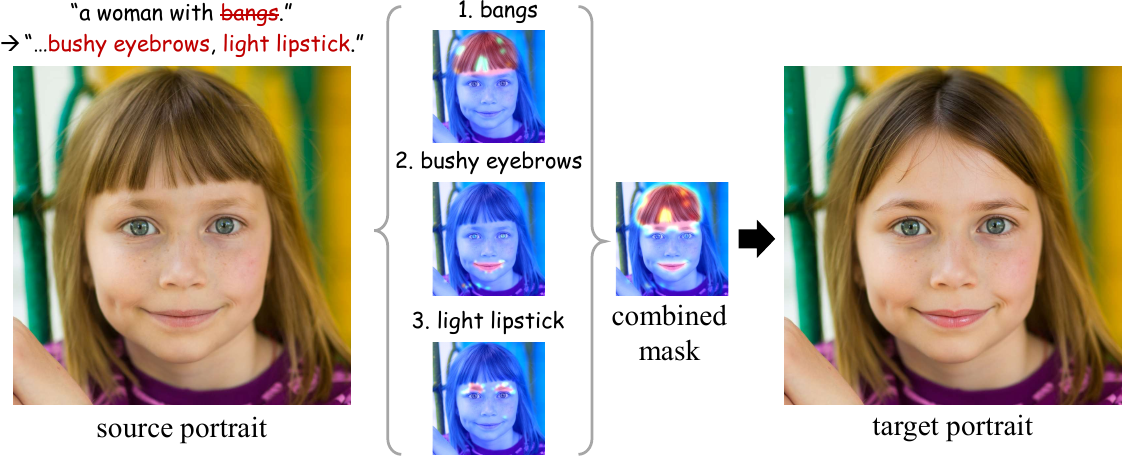}
    \caption{\textbf{Multi-attribute text-driven portrait editing.} Through a combined mask, Flux-Sculptor can remove bangs, apply lipstick, and thicken eyebrows within a single denoising process.}
    \label{fig:multi_attr_edit}
\end{figure}

    In the previous section, we primarily demonstrated single-attribute portrait editing. In practical applications, users may want to simultaneously modify multiple facial attributes for efficiency. Our Flux-Sculptor can be easily adapted to this task. The multi-attribute editing pipeline is shown in Figure~\ref{fig:multi_attr_edit}. We will split the input prompts into several single-attribute sub-prompts and generate their corresponding masks through PASL. Then, we combine these masks and input the combined mask and multi-attribute editing prompts into S2D-EC. In this way, we can achieve satisfying multi-attribute portrait editing in one turn.

\section{More Visualization Results}
\begin{figure*}
    \centering
    \includegraphics[width=\linewidth]{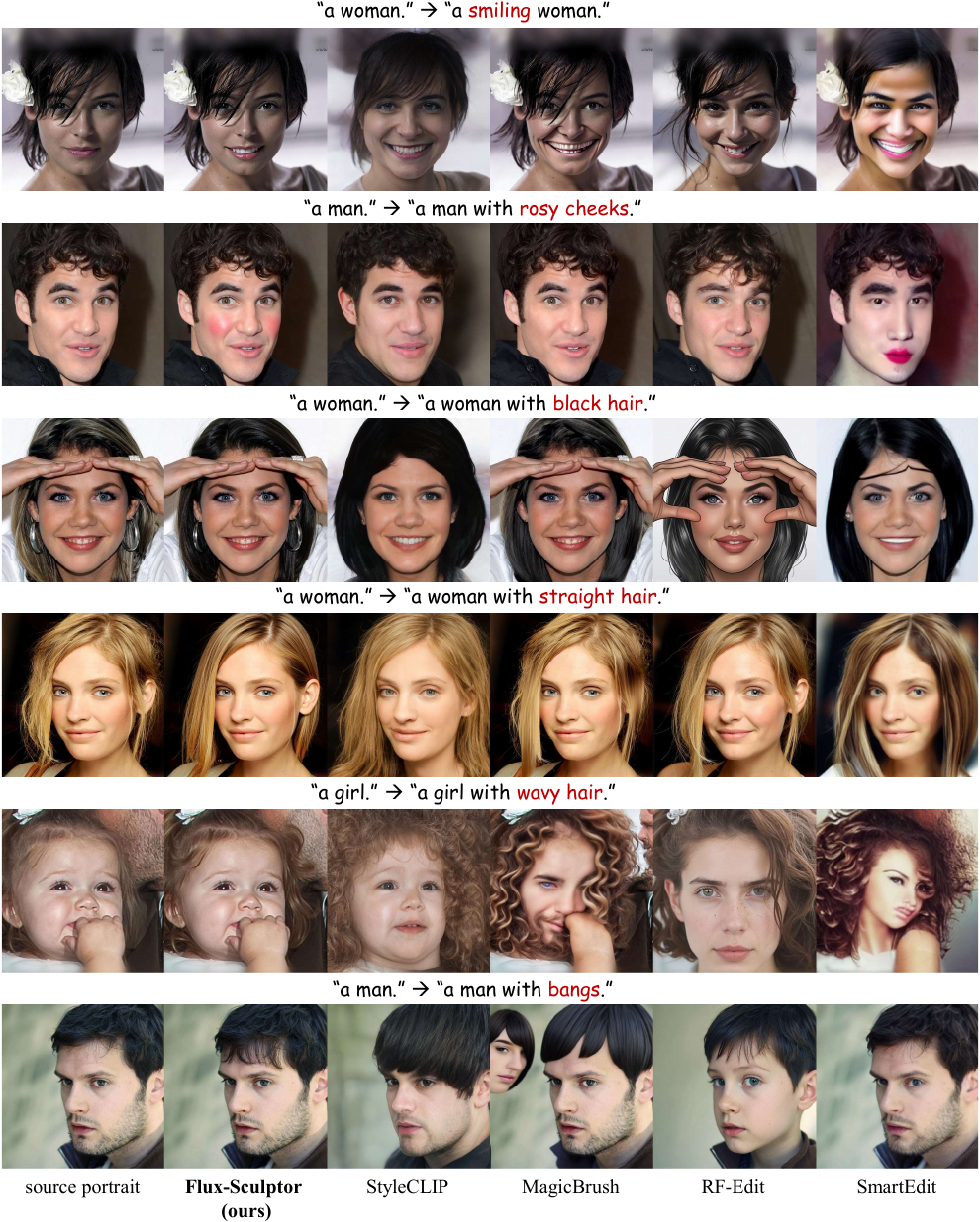}
    \caption{\textbf{More portrait editing visualization results.} We show the text-driven portrait editing results of our Flux-sculptor, StyleCLIP, MagicBrush, RF-Edit, and SmartEdit.}
    \label{fig:final_visualize_1}
\end{figure*}

\begin{figure*}
    \centering
    \includegraphics[width=\linewidth]{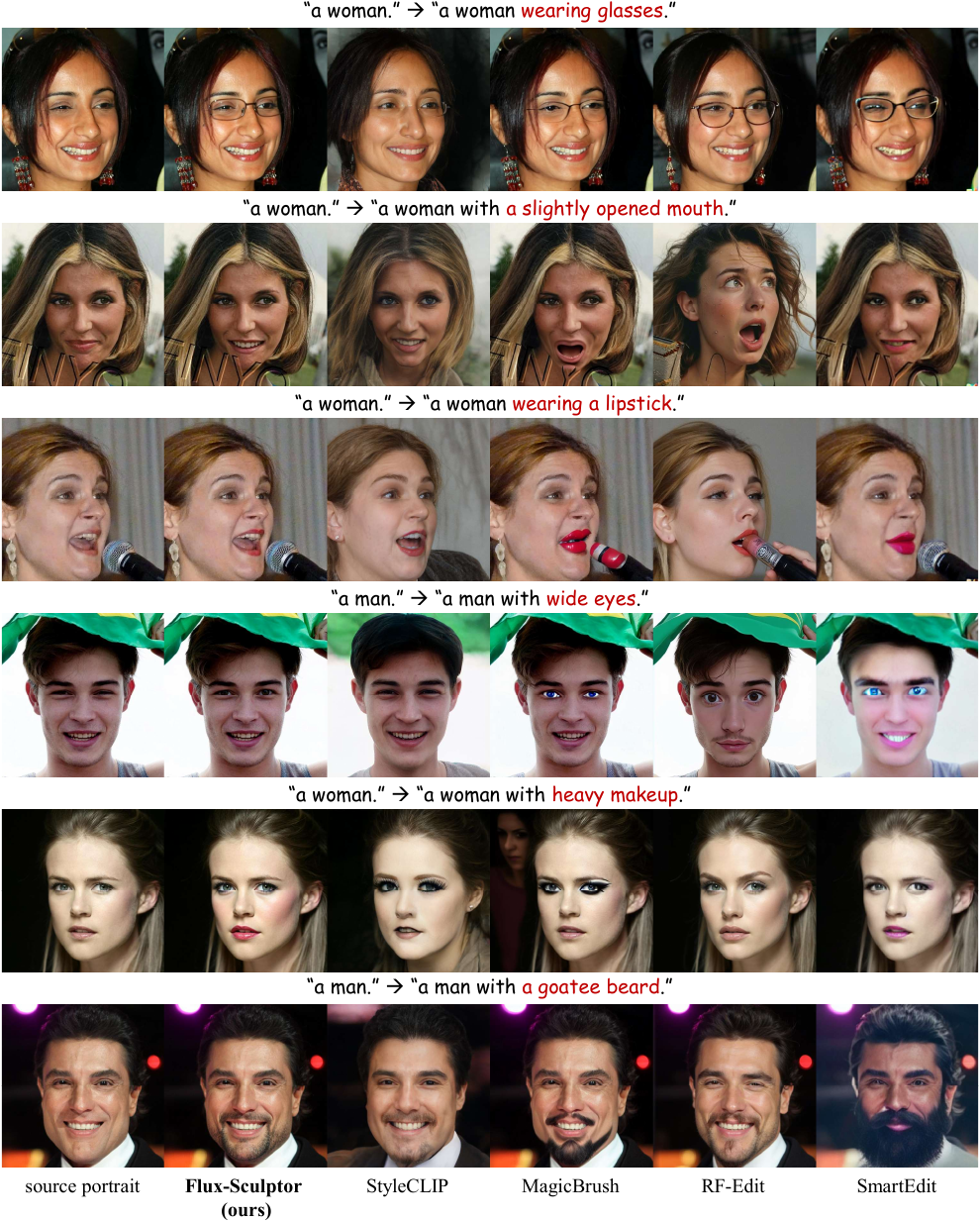}
    \caption{\textbf{More portrait editing visualization results.} We show the text-driven portrait editing results of our Flux-sculptor, StyleCLIP, MagicBrush, RF-Edit, and SmartEdit.}
    \label{fig:final_visualize_2}
\end{figure*}

    We visualize more text-driven portrait editing results of Flux-Sculptor and other state-of-the-art competitors in Figure~\ref{fig:final_visualize_1} and Figure~\ref{fig:final_visualize_2}. Extensive qualitative results show the superiority of our method.

%% file: main.bbl
\begin{thebibliography}{49}
\providecommand{\natexlab}[1]{#1}
\providecommand{\url}[1]{\texttt{#1}}
\expandafter\ifx\csname urlstyle\endcsname\relax
  \providecommand{\doi}[1]{doi: #1}\else
  \providecommand{\doi}{doi: \begingroup \urlstyle{rm}\Url}\fi

\bibitem[Adobe({\natexlab{a}})]{lightroom}
Adobe.
\newblock Lightroom.
\newblock lightroom.adobe.com/, {\natexlab{a}}.
\newblock 2025.3.5.

\bibitem[Adobe({\natexlab{b}})]{photoshop}
Adobe.
\newblock Photoshop.
\newblock www.adobe.com/products/ photoshop.html, {\natexlab{b}}.
\newblock 2025.3.5.

\bibitem[Brooks et~al.(2023)Brooks, Holynski, and Efros]{brooks2023instructpix2pix}
Tim Brooks, Aleksander Holynski, and Alexei~A Efros.
\newblock Instructpix2pix: Learning to follow image editing instructions.
\newblock In \emph{Proceedings of the IEEE/CVF conference on computer vision and pattern recognition}, pages 18392--18402, 2023.

\bibitem[Brown et~al.(2020)Brown, Mann, Ryder, Subbiah, Kaplan, Dhariwal, Neelakantan, Shyam, Sastry, Askell, et~al.]{brown2020language}
Tom Brown, Benjamin Mann, Nick Ryder, Melanie Subbiah, Jared~D Kaplan, Prafulla Dhariwal, Arvind Neelakantan, Pranav Shyam, Girish Sastry, Amanda Askell, et~al.
\newblock Language models are few-shot learners.
\newblock \emph{Advances in neural information processing systems}, 33:\penalty0 1877--1901, 2020.

\bibitem[Cao et~al.(2023)Cao, Wang, Qi, Shan, Qie, and Zheng]{cao2023masactrl}
Mingdeng Cao, Xintao Wang, Zhongang Qi, Ying Shan, Xiaohu Qie, and Yinqiang Zheng.
\newblock Masactrl: Tuning-free mutual self-attention control for consistent image synthesis and editing.
\newblock In \emph{Proceedings of the IEEE/CVF international conference on computer vision}, pages 22560--22570, 2023.

\bibitem[Chen et~al.(2019)Chen, Ishii, Bater, Darrach, Liao, Huynh, Reh, Nellis, Kumar, and Ishii]{chen2019association}
Jonlin Chen, Masaru Ishii, Kristin~L Bater, Halley Darrach, David Liao, Pauline~P Huynh, Isabel~P Reh, Jason~C Nellis, Anisha~R Kumar, and Lisa~E Ishii.
\newblock Association between the use of social media and photograph editing applications, self-esteem, and cosmetic surgery acceptance.
\newblock \emph{JAMA facial plastic surgery}, 21\penalty0 (5):\penalty0 361--367, 2019.

\bibitem[Corneanu et~al.(2024)Corneanu, Gadde, and Martinez]{corneanu2024latentpaint}
Ciprian Corneanu, Raghudeep Gadde, and Aleix~M Martinez.
\newblock Latentpaint: Image inpainting in latent space with diffusion models.
\newblock In \emph{Proceedings of the IEEE/CVF winter conference on applications of computer vision}, pages 4334--4343, 2024.

\bibitem[Couairon et~al.(2022)Couairon, Verbeek, Schwenk, and Cord]{couairon2022diffedit}
Guillaume Couairon, Jakob Verbeek, Holger Schwenk, and Matthieu Cord.
\newblock Diffedit: Diffusion-based semantic image editing with mask guidance.
\newblock \emph{arXiv preprint arXiv:2210.11427}, 2022.

\bibitem[Deng et~al.(2019)Deng, Guo, Xue, and Zafeiriou]{deng2019arcface}
Jiankang Deng, Jia Guo, Niannan Xue, and Stefanos Zafeiriou.
\newblock Arcface: Additive angular margin loss for deep face recognition.
\newblock In \emph{Proceedings of the IEEE/CVF conference on computer vision and pattern recognition}, pages 4690--4699, 2019.

\bibitem[Gao et~al.(2021)Gao, Wei, Bao, Gu, Chen, Wen, and Lian]{gao2021high}
Yue Gao, Fangyun Wei, Jianmin Bao, Shuyang Gu, Dong Chen, Fang Wen, and Zhouhui Lian.
\newblock High-fidelity and arbitrary face editing.
\newblock In \emph{Proceedings of the IEEE/CVF conference on computer vision and pattern recognition}, pages 16115--16124, 2021.

\bibitem[Gu et~al.(2019)Gu, Bao, Yang, Chen, Wen, and Yuan]{gu2019mask}
Shuyang Gu, Jianmin Bao, Hao Yang, Dong Chen, Fang Wen, and Lu Yuan.
\newblock Mask-guided portrait editing with conditional gans.
\newblock In \emph{Proceedings of the IEEE/CVF conference on computer vision and pattern recognition}, pages 3436--3445, 2019.

\bibitem[He et~al.(2017)He, Wang, Fu, Feng, Jiang, and Xue]{he2017adaptively}
Keke He, Zhanxiong Wang, Yanwei Fu, Rui Feng, Yu-Gang Jiang, and Xiangyang Xue.
\newblock Adaptively weighted multi-task deep network for person attribute classification.
\newblock In \emph{Proceedings of the 25th ACM international conference on Multimedia}, pages 1636--1644, 2017.

\bibitem[Hertz et~al.(2022)Hertz, Mokady, Tenenbaum, Aberman, Pritch, and Cohen-Or]{hertz2022prompt}
Amir Hertz, Ron Mokady, Jay Tenenbaum, Kfir Aberman, Yael Pritch, and Daniel Cohen-Or.
\newblock Prompt-to-prompt image editing with cross attention control.
\newblock \emph{arXiv preprint arXiv:2208.01626}, 2022.

\bibitem[Ho et~al.(2020)Ho, Jain, and Abbeel]{ho2020denoising}
Jonathan Ho, Ajay Jain, and Pieter Abbeel.
\newblock Denoising diffusion probabilistic models.
\newblock \emph{Advances in neural information processing systems}, 33:\penalty0 6840--6851, 2020.

\bibitem[Huang et~al.(2024)Huang, Xie, Wang, Yuan, Cun, Ge, Zhou, Dong, Huang, Zhang, et~al.]{huang2024smartedit}
Yuzhou Huang, Liangbin Xie, Xintao Wang, Ziyang Yuan, Xiaodong Cun, Yixiao Ge, Jiantao Zhou, Chao Dong, Rui Huang, Ruimao Zhang, et~al.
\newblock Smartedit: Exploring complex instruction-based image editing with multimodal large language models.
\newblock In \emph{Proceedings of the IEEE/CVF Conference on Computer Vision and Pattern Recognition}, pages 8362--8371, 2024.

\bibitem[Huynh-Thu and Ghanbari(2008)]{huynh2008scope}
Quan Huynh-Thu and Mohammed Ghanbari.
\newblock Scope of validity of psnr in image/video quality assessment.
\newblock \emph{Electronics letters}, 44\penalty0 (13):\penalty0 800--801, 2008.

\bibitem[Jiang et~al.(2024)Jiang, Huang, Xie, Xue, Liu, Wu, and Wong]{jiang2024hunting}
Le Jiang, Yan Huang, Lianxin Xie, Wen Xue, Cheng Liu, Si Wu, and Hau-San Wong.
\newblock Hunting blemishes: Language-guided high-fidelity face retouching transformer with limited paired data.
\newblock In \emph{Proceedings of the 32nd ACM International Conference on Multimedia}, pages 5102--5111, 2024.

\bibitem[Jiang et~al.(2021)Jiang, Huang, Pan, Loy, and Liu]{jiang2021talk}
Yuming Jiang, Ziqi Huang, Xingang Pan, Chen~Change Loy, and Ziwei Liu.
\newblock Talk-to-edit: Fine-grained facial editing via dialog.
\newblock In \emph{Proceedings of the IEEE/CVF International Conference on Computer Vision}, pages 13799--13808, 2021.

\bibitem[Jin et~al.(2024)Jin, Chen, Jin, Chen, Shi, Zheng, Zhu, and Ni]{jin2024toward}
Qiaoqiao Jin, Xuanhong Chen, Meiguang Jin, Ying Chen, Rui Shi, Yucheng Zheng, Yupeng Zhu, and Bingbing Ni.
\newblock Toward tiny and high-quality facial makeup with data amplify learning.
\newblock In \emph{European Conference on Computer Vision}, pages 340--356. Springer, 2024.

\bibitem[Karras et~al.(2017)Karras, Aila, Laine, and Lehtinen]{karras2017progressive}
Tero Karras, Timo Aila, Samuli Laine, and Jaakko Lehtinen.
\newblock Progressive growing of gans for improved quality, stability, and variation.
\newblock \emph{arXiv preprint arXiv:1710.10196}, 2017.

\bibitem[Karras et~al.(2019)Karras, Laine, and Aila]{karras2019style}
Tero Karras, Samuli Laine, and Timo Aila.
\newblock A style-based generator architecture for generative adversarial networks.
\newblock In \emph{Proceedings of the IEEE/CVF conference on computer vision and pattern recognition}, pages 4401--4410, 2019.

\bibitem[Kawar et~al.(2023)Kawar, Zada, Lang, Tov, Chang, Dekel, Mosseri, and Irani]{kawar2023imagic}
Bahjat Kawar, Shiran Zada, Oran Lang, Omer Tov, Huiwen Chang, Tali Dekel, Inbar Mosseri, and Michal Irani.
\newblock Imagic: Text-based real image editing with diffusion models.
\newblock In \emph{Proceedings of the IEEE/CVF conference on computer vision and pattern recognition}, pages 6007--6017, 2023.

\bibitem[Kim et~al.(2022)Kim, Kwon, and Ye]{kim2022diffusionclip}
Gwanghyun Kim, Taesung Kwon, and Jong~Chul Ye.
\newblock Diffusionclip: Text-guided diffusion models for robust image manipulation.
\newblock In \emph{Proceedings of the IEEE/CVF conference on computer vision and pattern recognition}, pages 2426--2435, 2022.

\bibitem[Kingma et~al.(2013)Kingma, Welling, et~al.]{kingma2013auto}
Diederik~P Kingma, Max Welling, et~al.
\newblock Auto-encoding variational bayes, 2013.

\bibitem[Kirillov et~al.(2023)Kirillov, Mintun, Ravi, Mao, Rolland, Gustafson, Xiao, Whitehead, Berg, Lo, Doll{\'a}r, and Girshick]{kirillov2023segany}
Alexander Kirillov, Eric Mintun, Nikhila Ravi, Hanzi Mao, Chloe Rolland, Laura Gustafson, Tete Xiao, Spencer Whitehead, Alexander~C. Berg, Wan-Yen Lo, Piotr Doll{\'a}r, and Ross Girshick.
\newblock Segment anything.
\newblock \emph{arXiv:2304.02643}, 2023.

\bibitem[Labs(2024)]{flux2024}
Black~Forest Labs.
\newblock Flux.
\newblock \url{https://github.com/black-forest-labs/flux}, 2024.

\bibitem[Li et~al.(2018)Li, Qian, Dong, Liu, Yan, Zhu, and Lin]{li2018beautygan}
Tingting Li, Ruihe Qian, Chao Dong, Si Liu, Qiong Yan, Wenwu Zhu, and Liang Lin.
\newblock Beautygan: Instance-level facial makeup transfer with deep generative adversarial network.
\newblock In \emph{Proceedings of the 26th ACM international conference on Multimedia}, pages 645--653, 2018.

\bibitem[Liu et~al.(2022)Liu, Gong, and Liu]{liu2022flow}
Xingchao Liu, Chengyue Gong, and Qiang Liu.
\newblock Flow straight and fast: Learning to generate and transfer data with rectified flow.
\newblock \emph{arXiv preprint arXiv:2209.03003}, 2022.

\bibitem[Loshchilov et~al.(2017)Loshchilov, Hutter, et~al.]{loshchilov2017fixing}
Ilya Loshchilov, Frank Hutter, et~al.
\newblock Fixing weight decay regularization in adam.
\newblock \emph{arXiv preprint arXiv:1711.05101}, 5:\penalty0 5, 2017.

\bibitem[Lugmayr et~al.(2022)Lugmayr, Danelljan, Romero, Yu, Timofte, and Van~Gool]{lugmayr2022repaint}
Andreas Lugmayr, Martin Danelljan, Andres Romero, Fisher Yu, Radu Timofte, and Luc Van~Gool.
\newblock Repaint: Inpainting using denoising diffusion probabilistic models.
\newblock In \emph{Proceedings of the IEEE/CVF conference on computer vision and pattern recognition}, pages 11461--11471, 2022.

\bibitem[Pang et~al.(2023)Pang, Zhang, Quan, Fan, Cun, Shan, and Yan]{pang2023dpe}
Youxin Pang, Yong Zhang, Weize Quan, Yanbo Fan, Xiaodong Cun, Ying Shan, and Dong-ming Yan.
\newblock Dpe: Disentanglement of pose and expression for general video portrait editing.
\newblock In \emph{Proceedings of the IEEE/CVF Conference on Computer Vision and Pattern Recognition}, pages 427--436, 2023.

\bibitem[Patashnik et~al.(2021)Patashnik, Wu, Shechtman, Cohen-Or, and Lischinski]{patashnik2021styleclip}
Or Patashnik, Zongze Wu, Eli Shechtman, Daniel Cohen-Or, and Dani Lischinski.
\newblock Styleclip: Text-driven manipulation of stylegan imagery.
\newblock In \emph{Proceedings of the IEEE/CVF international conference on computer vision}, pages 2085--2094, 2021.

\bibitem[Peebles and Xie(2023)]{peebles2023scalable}
William Peebles and Saining Xie.
\newblock Scalable diffusion models with transformers.
\newblock In \emph{Proceedings of the IEEE/CVF international conference on computer vision}, pages 4195--4205, 2023.

\bibitem[Preechakul et~al.(2022)Preechakul, Chatthee, Wizadwongsa, and Suwajanakorn]{preechakul2022diffusion}
Konpat Preechakul, Nattanat Chatthee, Suttisak Wizadwongsa, and Supasorn Suwajanakorn.
\newblock Diffusion autoencoders: Toward a meaningful and decodable representation.
\newblock In \emph{Proceedings of the IEEE/CVF conference on computer vision and pattern recognition}, pages 10619--10629, 2022.

\bibitem[Radford et~al.(2021)Radford, Kim, Hallacy, Ramesh, Goh, Agarwal, Sastry, Askell, Mishkin, Clark, et~al.]{radford2021learning}
Alec Radford, Jong~Wook Kim, Chris Hallacy, Aditya Ramesh, Gabriel Goh, Sandhini Agarwal, Girish Sastry, Amanda Askell, Pamela Mishkin, Jack Clark, et~al.
\newblock Learning transferable visual models from natural language supervision.
\newblock In \emph{International conference on machine learning}, pages 8748--8763. PmLR, 2021.

\bibitem[Ravi et~al.(2024)Ravi, Gabeur, Hu, Hu, Ryali, Ma, Khedr, R{\"a}dle, Rolland, Gustafson, Mintun, Pan, Alwala, Carion, Wu, Girshick, Doll{\'a}r, and Feichtenhofer]{ravi2024sam2}
Nikhila Ravi, Valentin Gabeur, Yuan-Ting Hu, Ronghang Hu, Chaitanya Ryali, Tengyu Ma, Haitham Khedr, Roman R{\"a}dle, Chloe Rolland, Laura Gustafson, Eric Mintun, Junting Pan, Kalyan~Vasudev Alwala, Nicolas Carion, Chao-Yuan Wu, Ross Girshick, Piotr Doll{\'a}r, and Christoph Feichtenhofer.
\newblock Sam 2: Segment anything in images and videos.
\newblock \emph{arXiv preprint arXiv:2408.00714}, 2024.

\bibitem[Rombach et~al.(2022)Rombach, Blattmann, Lorenz, Esser, and Ommer]{rombach2022high}
Robin Rombach, Andreas Blattmann, Dominik Lorenz, Patrick Esser, and Bj{\"o}rn Ommer.
\newblock High-resolution image synthesis with latent diffusion models.
\newblock In \emph{Proceedings of the IEEE/CVF conference on computer vision and pattern recognition}, pages 10684--10695, 2022.

\bibitem[Rout et~al.(2024)Rout, Chen, Ruiz, Caramanis, Shakkottai, and Chu]{rout2024semantic}
Litu Rout, Yujia Chen, Nataniel Ruiz, Constantine Caramanis, Sanjay Shakkottai, and Wen-Sheng Chu.
\newblock Semantic image inversion and editing using rectified stochastic differential equations.
\newblock \emph{arXiv preprint arXiv:2410.10792}, 2024.

\bibitem[Si et~al.(2024)Si, Huang, Jiang, and Liu]{si2024freeu}
Chenyang Si, Ziqi Huang, Yuming Jiang, and Ziwei Liu.
\newblock Freeu: Free lunch in diffusion u-net.
\newblock In \emph{Proceedings of the IEEE/CVF Conference on Computer Vision and Pattern Recognition}, pages 4733--4743, 2024.

\bibitem[Song et~al.(2020)Song, Meng, and Ermon]{song2020denoising}
Jiaming Song, Chenlin Meng, and Stefano Ermon.
\newblock Denoising diffusion implicit models.
\newblock \emph{arXiv preprint arXiv:2010.02502}, 2020.

\bibitem[Sudre et~al.(2017)Sudre, Li, Vercauteren, Ourselin, and Jorge~Cardoso]{sudre2017generalised}
Carole~H Sudre, Wenqi Li, Tom Vercauteren, Sebastien Ourselin, and M Jorge~Cardoso.
\newblock Generalised dice overlap as a deep learning loss function for highly unbalanced segmentations.
\newblock In \emph{Deep Learning in Medical Image Analysis and Multimodal Learning for Clinical Decision Support: Third International Workshop, DLMIA 2017, and 7th International Workshop, ML-CDS 2017, Held in Conjunction with MICCAI 2017, Qu{\'e}bec City, QC, Canada, September 14, Proceedings 3}, pages 240--248. Springer, 2017.

\bibitem[Sun et~al.(2024)Sun, Fang, Wu, Zhang, Zang, Kong, Xiong, Lin, and Wang]{sun2024alpha}
Zeyi Sun, Ye Fang, Tong Wu, Pan Zhang, Yuhang Zang, Shu Kong, Yuanjun Xiong, Dahua Lin, and Jiaqi Wang.
\newblock Alpha-clip: A clip model focusing on wherever you want.
\newblock In \emph{Proceedings of the IEEE/CVF conference on computer vision and pattern recognition}, pages 13019--13029, 2024.

\bibitem[Wang et~al.(2024{\natexlab{a}})Wang, Lin, del Molino, Wang, Feng, and Shen]{wang2024maniclip}
Hao Wang, Guosheng Lin, Ana~Garc{\'\i}a del Molino, Anran Wang, Jiashi Feng, and Zhiqi Shen.
\newblock Maniclip: Multi-attribute face manipulation from text.
\newblock \emph{International Journal of Computer Vision}, 132\penalty0 (10):\penalty0 4616--4632, 2024{\natexlab{a}}.

\bibitem[Wang et~al.(2024{\natexlab{b}})Wang, Pu, Qi, Guo, Ma, Huang, Chen, Li, and Shan]{wang2024taming}
Jiangshan Wang, Junfu Pu, Zhongang Qi, Jiayi Guo, Yue Ma, Nisha Huang, Yuxin Chen, Xiu Li, and Ying Shan.
\newblock Taming rectified flow for inversion and editing.
\newblock \emph{arXiv preprint arXiv:2411.04746}, 2024{\natexlab{b}}.

\bibitem[Wang et~al.(2004)Wang, Bovik, Sheikh, and Simoncelli]{wang2004image}
Zhou Wang, Alan~C Bovik, Hamid~R Sheikh, and Eero~P Simoncelli.
\newblock Image quality assessment: from error visibility to structural similarity.
\newblock \emph{IEEE transactions on image processing}, 13\penalty0 (4):\penalty0 600--612, 2004.

\bibitem[Yue et~al.(2023)Yue, Guo, Ning, Cui, Zhu, and Yuan]{yue2023chatface}
Dongxu Yue, Qin Guo, Munan Ning, Jiaxi Cui, Yuesheng Zhu, and Li Yuan.
\newblock Chatface: Chat-guided real face editing via diffusion latent space manipulation.
\newblock \emph{arXiv preprint arXiv:2305.14742}, 2023.

\bibitem[Zhang et~al.(2023)Zhang, Mo, Chen, Sun, and Su]{zhang2023magicbrush}
Kai Zhang, Lingbo Mo, Wenhu Chen, Huan Sun, and Yu Su.
\newblock Magicbrush: A manually annotated dataset for instruction-guided image editing.
\newblock \emph{Advances in Neural Information Processing Systems}, 36:\penalty0 31428--31449, 2023.

\bibitem[Zhang et~al.(2018)Zhang, Isola, Efros, Shechtman, and Wang]{zhang2018unreasonable}
Richard Zhang, Phillip Isola, Alexei~A Efros, Eli Shechtman, and Oliver Wang.
\newblock The unreasonable effectiveness of deep features as a perceptual metric.
\newblock In \emph{Proceedings of the IEEE conference on computer vision and pattern recognition}, pages 586--595, 2018.

\bibitem[Zhang et~al.(2024)Zhang, Zhang, Song, Zhang, Tang, and Liu]{zhang2024stable}
Yuxuan Zhang, Qing Zhang, Yiren Song, Jichao Zhang, Hao Tang, and Jiaming Liu.
\newblock Stable-hair: Real-world hair transfer via diffusion model.
\newblock \emph{arXiv preprint arXiv:2407.14078}, 2024.

\end{thebibliography}
